\documentclass[sigconf]{acmart}

\usepackage{graphicx}
\usepackage{booktabs} 
\usepackage{float}
\usepackage{amsmath}
\usepackage{bbm}
\usepackage{subcaption}
\usepackage{slashbox}
\usepackage{hyperref} 
\usepackage[vertfit]{breakurl}
\usepackage{pbox}
\usepackage{paralist}
\usepackage[]{color}
\usepackage{algorithm}
\usepackage[noend]{algpseudocode}
\usepackage{comment}
\usepackage[bottom]{footmisc}
\usepackage{arydshln}
\usepackage{pifont}
\newcommand{\cmark}{\ding{51}}%
\newcommand{\xmark}{\ding{55}}%
\newcommand{\mmark}{{\small\ding{51}\hspace{-.2cm}\ding{55}}}
\DeclareMathOperator*{\argmin}{arg\,min}
\DeclareMathOperator*{\argmax}{arg\,max}

\usepackage{multirow}
\usepackage{balance}
\usepackage{epstopdf}
\usepackage{wrapfig}

\usepackage{tikz}
\usetikzlibrary{arrows, decorations.markings}
\colorlet{mygray}{black!60}
\tikzset{thicker line small arrows m/.style args={#1in#2}{
		draw=#2,
		solid,
		line width=#1,
		shorten >=1mm,
		decoration={
			markings,
			mark=at position 1.0 with {\arrow[fill=#2,thin]{triangle 90}}
		},
		postaction={decorate}
}}

\newcommand{\add}[1]{\textcolor{black}{{#1}}} 

\newtheorem{problem}{Problem}

\newcommand{\TechReport}{}
\newcommand{\available}[2]{{#1}^{a}_{{#2}}}
\newcommand{\selected}[2]{{#1}_{{#2}}}
\newcommand{\samples}{X}
\newcommand{\labels}{Y}

\begin{document}

\title{From Limited Annotated Raw Material Data to Quality Production Data: A Case Study in the Milk Industry \\(Technical Report)}
\subtitle{Don’t Cry Over Spilt Milk}
\author{Roee Shraga}
\email{shraga89@campus.technion.ac.il}
\affiliation{%
\institution{Technion, Haifa, Israel}
\city{}
\country{}
}
\author{Gil Katz}
\email{gil@afimilk.co.il}
\affiliation{%
	\institution{Afimilk, Afikim, Israel}
	\city{}
	\country{}
}
\author{Yael Badian, Nitay Calderon, Avigdor Gal}
\email{{yaelbadian,nitay}@campus.technion.ac.il}
\email{avigal@technion.ac.il}
\affiliation{%
\institution{Technion, Haifa, Israel}
\city{}
\country{}
}

%

\begin{abstract}
	

Industry 4.0 offers opportunities to combine multiple sensor data sources using IoT technologies for better utilization of raw material in production lines. A common belief that data is readily available (the big data phenomenon), is oftentimes challenged by the need to effectively acquire quality data under severe constraints. In this paper we propose a design methodology, using active learning to enhance learning capabilities, for building a model of production outcome using a constrained amount of raw material training data. The proposed methodology extends existing active learning methods to effectively solve regression-based learning problems and may serve settings where data acquisition requires excessive resources in the physical world. We further suggest a set of qualitative measures to analyze learners performance. The proposed methodology is demonstrated using an actual application in the milk industry, where milk is gathered from multiple small milk farms and brought to a dairy production plant to be processed into cottage cheese. 
\end{abstract}

\maketitle

\section{Introduction}
The evolution of data acquisition, management, analytics, and visualization has led to coining the term {\em big data}. Big data encompasses emerging technological advancements such as IoT -- Internet of things (acquisition), cloud computing (management), and machine learning (analytics), and providing an exciting arena for new and challenging research agenda. 
Big data is commonly characterized via a set of ``V"s, out of which {\em volume} characterizes the need to gather, manage, and analyze massive amounts of data. A major source of data is the one stemming from sensors, driven by IoT technology that offers standards for bringing multiple sources of data together to be stored, analyzed, and visualized.

Industry 4.0 is as an encompassing term to sensor data deployment using IoT technologies in factories where machines augmented with wireless connectivity and sensors are connected to monitor an entire production line for autonomous decision making. 

Industry 4.0 faces challenges that involve
utilizing big data technologies and analytics at scale. To revolutionize supply chain management, there is a need to handle the accelerating scale, scope and depth of data.
The end goal is to enable complex supplier networks to optimize the flow of products, information, and finances. Such optimization allows companies to improve their overall business efficiency, while hand-in-hand personalize their services and tailoring them to customer needs, thus creating sustainable value. 

Modern supply chain companies run sophisticated workflows (manufacturing, transportation, sales, logistics, {\em etc.}) and make use of IoT technology to receive streams of data that flood in from suppliers, customers, machinery, environment, and delivery channels. For example, sensor-equipped machines offer raw material readings that may be used for anticipating the performance of raw material in a production line. One of the toughest contemporary obstacles involve 
the need to integrate multiple data sources on a massive scale to capitalize on this rich pool of data and create invaluable knowledge and insights~\cite{yao2014edal} (big data {\em variety}). 

In this work we tackle a problem of creating a training dataset for controlled learning of the impact of raw material on a production line under severe limitations of data acquisition. Despite the generous availability of sensor data both on the side of the raw materials and on the side of the production line, it is still a challenge to construct a training dataset from existing data mainly due to the integration problem. While the raw material may be monitored carefully, once it is being shipped to the factory it may be combined with other raw material and then spread out to serve in multiple production lines, all of which may lack any control and monitoring. To allow a controlled setting, the raw material provider and the manufacturer need to agree on a protocol of providing raw material of a certain quality to be processed in a specific production line, so as to keep as many of the variables that may affect the production line in a controlled state. 

We propose an active learning solution to effectively acquire IoT data for quality training purposes.\footnote{This paper is the technical report of~\cite{shraga2021limited}.} Our contribution is threefold:
\begin{compactitem}
	\item \textbf{Modeling}: We propose a model for unifying the representation of raw material's IoT sensor data in a production line, which lends itself well to (active) learning tasks.
	\item \textbf{Algorithmic}: We extend active learning state-of-the-art to effectively solve regression-based problems and suggest qualitative measures to analyze learners performance.
	\item \textbf{Application}: We provide a detailed dairy production industry case study to illustrate the potential of active learning to creating datasets that adhere to the application's need. 
\end{compactitem}
Using a controlled experiment, we demonstrate the effectiveness of applying active learning to generate a dataset for testing the impact of milk features on the production of cottage cheese.
The rest of the paper is organized as follows. We start with an illustrative case study from the milk industry, demonstrating the need and possible solution to the creation of a training set using a small amount of sample collection (Section~\ref{sec:useCase}). A model for integrated sensor systems and a problem definition are given in Section~\ref{sec:modelProblem}. Section~\ref{sec:alg} offers our solution to effectively learn the impact of raw material on production outcome. An empirical evaluation (Section~\ref{sec:exp}) is followed by related work (Section~\ref{sec:related}) and a concluding discussion (Section~\ref{sec:conclusions}). 
\section{Case Study: the Milk Industry}
\label{sec:useCase}
In this section we provide a specific use case of the utilization of IoT in the food industry. We focus on the milk industry and start our presentation with a description of the farm setting (Section~\ref{sec:Milk farm's perspective}), followed by the dairy setting (Section~\ref{sec:dairy}). We conclude with a description of the available sensor and auxiliary data (Section~\ref{sec:useCaseData}).
\subsection{Milk Farm's Perspective}
\label{sec:Milk farm's perspective}
Milk has played an important role in mammals evolution, as the most effective nutrient, contributing to survival success of these mammals descendants. Humans exercised lactation control in mammals, for nutritional purposes, since the beginning of civilization~\cite{Blackburn1989} and bovine milk is the major raw material in the dairy industry.
 
Cows must calve to produce milk and a lactation cycle is a period between one calving to the next. This cycle is split into four phases, namely early, mid, late lactation of actual milk production (each approximately 100 days) and a dry period (approximately 65 days). The next calving begins once the dry period is over.
  
To produce milk during these four stages, cows forgo physiological changes from mobilizing body reserves at the beginning of the cycle through body reserves regain and finally rumen rehabilitation before next calving~\cite{Lewis1998, Andersen2001}. Dairy farmers objective is to obtain an optimal control of the lactation cycle by planning cow performance that maximizes the farm income~\cite{Erdman2000}.
 

Over the last century, the production of dairy cattle milk was increased enormously, mainly due to the development of quantitative population genetics~\cite{Weller2007}. The modern high producing dairy cows are a product of genetic selection for increased milk production as a primary trait~\cite{Shook2006, Rosa2015}. 
A modern milk production procedure takes into account the energy transfer from digestion of feed, tissue metabolism and muscle activity of cows~\cite{Maltz2013}. A herd (group of cows) is managed using feed and reproduction strategies, as well as health control through the lactation cycle.

The basis for farmers' decision making is the energy balance of the individual cow. However, the production of a farm is an aggregated effort of the whole herd. At any given time, the cows in the herd differ in their lactation stage, physiological condition, health status {\em etc.}, all of which have direct implications on the bulk milk, which is composed of all individual milked cows.


Milk production enjoys economy of scale. Therefore, a dominating trend in global dairy industry involves a drop in the number of herds along an increase in the herds sizes. Owner operated farms are evolving into big organizations~\cite{Kinsel2013}, small family farms evolve into large dairies with thousands of animals, and traditional $10$-$50$ cows family farms built in populated areas are replaced with large dairies in agricultural areas, with thousands of milking cows.


While intimate familiarity with each individual animal is no longer an option in large dairies, the basic production unit remains the individual cow with its lactation cycle. Therefore, monitoring a heterogeneous collection of basic production units remains a necessity in this industry, leading to the creation of {\em precision dairy farming} (PDF). PDFs are computerized management systems, such as {\sf afifarm 5}, which was developed by our industrial partner Afimilk,\footnote{Afimilk LTD, Afikim, Israel, \url{https://www.afimilk.com/afifarm}} and uses information technology to manage the smallest production unit possible, expressing its genetic potential in accordance with the economical goal and animal well-being. {\sf Afifarm 5} uses a reliable ID system, a set of sensors to automatically collect relevant data and the farming know-how to perform analysis of the data to support productive decision making. IoT sensors provide detailed online data about each individual cow in a herd. With proper physiological labeling and interpretation, this data can be used to support the decision making in farms~\cite{Maltz2010, Ezra2016}.

\subsection{Milk Dairy's Perspective}
\label{sec:dairy}
Milk is a biological complex fluid. 
While the major component of milk is water or skim liquid, milk contains up to 14\% of total solids, double than most vegetables.  
Milk contains fat, protein, lactose, minerals, and more than 250 chemical compounds and metabolites such as
micronutrients ({\em e.g.}, vitamins), amino acids, trace minerals, hormones, and enzymes~\cite{Jensen2002, 14517}. 
The rich content of milk allows it to be manufactured into other, stable dairy products such as butter, cheese, dried milks, ice cream, and condensed milk using complex and sensitive processes that have been refined over many generations.  
To date, behavior of complex fluids like milk, whose dynamics are highly nonlinear, is still not well understood and 
self assembly into different types of aggregates is complex~\cite{Avinoam1996, 14518}. Aiming to understand these dynamics per a required product is at the heart of the know-how of the dairy production line.

Consider {\em cheese}, a product of a formation of protein matrix originating from casein micelles held together by weak physical forces trapping fat globules within the matrix: ``The physical properties of cheese ({\em i.e.}, body/texture, melt/stretch, and color) are influenced by initial cheese[-]milk composition, manufacturing procedures, and maturation conditions. Two of the most important factors influencing these properties are the condition of the CN [(casein)] particles in cheese ({\em e.g.}, interactions between and within molecules, as well as the amount of Ca [(Calcium)] associated with these particles) and the extent of proteolysis. These are in turn influenced by various environmental conditions such as pH development, temperature, and ionic strength. Therefore, how individual CN molecules, or aggregates of many CN molecules interact, is vital in understanding the physical and chemical properties of cheese.''~\cite{Lucey2003}.

Complex fluid dynamics led to a setting where there is no uniform set of principles to fully control milk processing. Milk processed at the production line has low repeatability between days, production lines, and farms. Moreover, milk batches containing the same concentrations of solids, fat, protein, and even casein were shown to result in different cheese yields~\cite{Law2010}. Contemporary large scale dairy plants use inaccurate measures such as solids level (fat and protein), Somatic Cell Count, and bacteria counts (herd udder health) to quantify daily milk quality and to scale milk payment. It is this challenge we aim at tackling in this work, offering an effective algorithm for quality data acquisition using IoT data.

\subsection{From Farmer to Dairy}
\label{sec:useCaseData}
The farmer (Section~\ref{sec:Milk farm's perspective}) and dairy processors (Section~\ref{sec:dairy}) address two different qualitative evaluation of milk.
On the one hand, at the dairy, milk quality is measured on an industrial level and related to economics through product composition and suitability, as well as the production yield. Milk age and storage conditions influence milk quality as well~\cite{Barbano2000}. Finally, end user needs regarding good and bad fatty acids profiling should also be taken into account. On the other hand, farmer's milk quality, properties, and composition are addressed at the udder level and depend on breed, nutrition, lactation cycle, and udder infection~\cite{Leitner2008, Munro84}. The interaction between farmer and dairy occurs in the farm's bulk tank. The input from the farmer to the bulk tank is milk from selected individual cows (farmer's perspective). The output of the bulk tank at the dairy is a composed milk from individual milked cows (dairy's perspective).
  
This work attempts to remove the barrier between processor (dairy) and farmer, employing PDF data (Section~\ref{sec:Milk farm's perspective}). PDF data, collected from individual production units (cows), can be used to characterize bulk tank's properties and, when merged with production line data, to optimize production. This process is a data management challenge, handling data from multiple sources, of various types (including categorical, quantitative, subjective, and objective data), and at different time scales. To demonstrate the rich IoT data content of this automated process, we use {\sf afifarm 5} to extract the following daily data:
\begin{compactitem}
	\item Individual cow wearable (afiactII motion) sensors that measure cow behavior including motions, activity, lying time, and lying bouts. Each cow is equipped with a single sensor. 
	\item Milk data sensors using milk spectrometer and milk flowmeter that measure yield, flow, conductivity, milk components, and milk NIR spectra 2-5 times a day using afilab\textsuperscript{TM} milk analyzer and afimilk MPC\textsuperscript{TM} milk meter. We note that these sensors are subject to frequent malfunctions.
	\item A walk-through scale, measuring cow weight.
	\item Third party data such as health report (Veterinerian), feed (Nutritionist), and farmer's action reports. Such data suffers from low integrity and inconsistency across farms.
\end{compactitem}
IoT data is enriched with data from production lines (yield, process parameters) and chemistry laboratories (coagulation properties, proteins profiling, saturated and un-saturated fatty acids, cheese and yougurt production pilots, and electronic sniffer).


In this work, we present a methodology, based on active learning, to create a schedule for experimenting with IoT data and build a prediction model of raw milk's rennet activity during cheese production. 
Milk clotting activity may only be measured by the rapidity with which the enzyme clots milk under a set of specified conditions (RCT) or the Curd Firmness (CF)~\cite{rennet1988}. Such retrospective empirical evaluation of milk's coagulation properties provide limited means to assess milk quality. Yet, there is a need for predictive mechanisms to assist dairies in understanding the quality of milk that enters a manufacturing line. As the labeling quality measurement, our measure of choice is CF, measured with an Optigraph instrument (Ysebaert, Frepillon, France)~\cite{forma82, opti2005}. CF is measured in the lab in a process that takes 60 to 90 minutes for a batch of 10 samples.

\section{Model and Problem Definition}
\label{sec:modelProblem}
\begin{figure}[t]
	\centering
	\includegraphics[width=\linewidth, height=.6\linewidth]{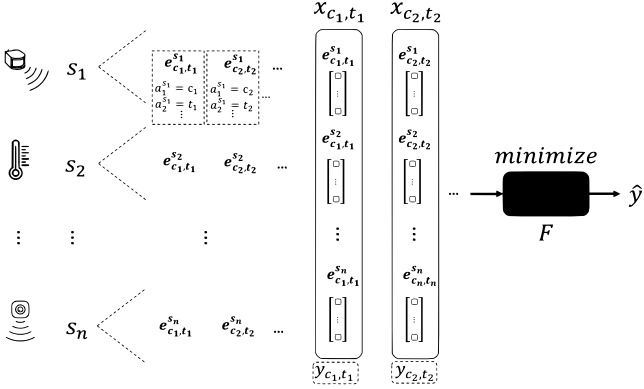}
	\caption{Farm Model Illustration}
	\label{fig:acro}
\end{figure}

In this section, we present a model of the data at the farm (Section~\ref{sec:model} \add{and Figure~\ref{fig:acro})} and use it to introduce the problem definition that serves as the focus of this work (Section~\ref{sec:problem}).
\subsection{Model}

\label{sec:model}
Let $S=\langle s_1, s_2, \dots, s_n\rangle$ be a set of sensors (ordered for convenience of presentation). We use a \emph{point-based semantics} for time and for simplicity sake we use $\mathbb{Z}^+$ as a time domain. Using $S$ and a discrete time domain, the data emitted by sensor $s\in S$ is modeled as a data stream $\sigma^s=\langle e_1^s, e_2^s, \ldots \rangle$, which contains a (possibly infinite) sequence of \emph{events}, 
with a similar structure and meaning. A sensor $s$ belongs to a sensor type, denoted by $s.type$, which determines the structure of events generated by $s$. Such a structure of an event is given as a set of attributes, $\mathcal{A}^s=\{a_1^s, a_2^s,\ldots, a_p^s\}$, with $dom(a_i^s)$ as the domain of attribute $a_i^s$ ($1\leq i\leq p$), dictated by its type. We denote by $dom(\mathcal{A}^s)=\times_{1\leq i\leq p}dom(a_i^s)$ the domain of sensor $s$, where $\times$ is the generalized Cartesian product.  We use $e.a^s$ to denote the value of attribute $a^s$ of event $e$ and $e.\mathcal{A}^s$ to denote the values assigned with event $e$. Finally, the joint domain of all sensor values is denoted as $dom(\mathcal{A})=\times_{s\in S}dom(\mathcal{A}^s)$.

In common to all events is an attribute $e.id$, which uniquely identifies the event. We also assume that each event $e$ is associated with a timestamp attribute $e.ts$, referring to the time it was emitted. We require each sensor to emit $e.cid$, explicitly creating the notion of a {\em case} to identify a unique identity that associates all events that share the same case id. Let $C$ be a set of $cid$ values and $c\in C$ a single case id. A {\em sample} of $S$ at time $t\in\mathbb{Z}^+$ for case $c$ is defined as $x_{c,t}=\langle e_{c,t}^{s_1}.\mathcal{A}^{s_1}, e_{c,t}^{s_2}.\mathcal{A}^{s_2},\dots,e_{c,t}^{s_n}.\mathcal{A}^{s_n}\rangle$, where $e_{c,t}$ stands for an event $e$ such that $e.cid=c$ and $e.ts=t$.


\begin{example}\label{ex:sensors}
\begin{sloppypar}
Let $S_{ex} =\langle s_{spectrometer}, s_{status}\rangle$ be an example of two sensors based on our case study (Section~\ref{sec:useCase}). The sensor $s_{spectrometer}$ represents the dynamics of milk coagulation process and composed of the attributes $\{fat, protein, lactose, IgG, casein, Oa, Sufa, Mufa\}$. $casein$ with domain $dom(casein) = [1.5,3]$, for example, represents $\kappa$‐casein hydrolysis~\cite{kappa93}, which is the primary phase of the enzymatic milk coagulation. Other attribute examples in $s_{spectromemeter}$ are $Oa$, $Sufa$, $Mufa$ measuring Oleaic Acid and Poly-unsaturated fatty acids and Mono-unsaturated fatty acids, respectively. 
$s_{status}$ tracks cows status with, among others, the attributes days in milk ($DIM$, $dom(DIM) = \{0,1,\dots, 305\}$) and $GYN$ (gynecological status).
\end{sloppypar}
\end{example}
%

In the context of a production line, we define an outcome measure $Y$ to be some assessment of performance over a sample, which takes its values from $dom(Y)$. Given a sample $x_{c,t}$, we mark by $y_{c,t}\in dom(Y)$ the measure assigned with it. 
$f:dom(\mathcal{A})\rightarrow dom(Y)$ maps sensor values to a measure.

\subsection{Problem Definition}
\label{sec:problem}
The process of collecting milk, producing cheese, and analyzing its quality is expensive, requiring many resources. Hence, number of samples available for learning is limited. Moreover, the dairy can only allocate limited facilities for the benefit of an experiment and therefore only a few samples can be processed each week. We therefore have a constrained setting with a limited time for experimenting (an {\em experiment horizon} $\{1,2,\ldots,T\}$) and a limited number of samples per iteration (a budget $b$).

We position this problem as an \emph{active learning}~\cite{settles1995active} problem, aiming at selecting the $b$ most informative samples for labeling throughout the experiment horizon $\{1,2,\ldots,T\}$.

A common setting for active learning assumes the availability of a static pool from which samples are drawn. The sample collection in the milk industry presents a unique challenge, in which we do not have such a privilege. In our scenario, we are presented with an infinite domain of samples $dom(\mathcal{A})$ from which, at each time $t$, we are given a selection of a different set of authorized samples $\available{\samples}{t}$. To be more concrete, each week our sample set changes based on the specific characteristics of milk that cows provide. The variation depends on external aspects ({\em e.g.}, weather, food provision) as well as internal ({\em e.g.}, cow health, days into milking cycle).

Our goal is to find a mapping from input samples in $dom(\mathcal{A})$ to an outcome measure $dom(Y)$, by selecting at time $t$ the best available samples $\selected{\samples}{t}$ out of $\available{\samples}{t}$, subject to a budget constraint. 


\begin{problem}\label{prob}
	Let $\{1,2,\ldots,T\}$ be an experiment horizon and {\em b} a sample budget. At time $t\leq T$, given $\available{\samples}{t}$ and knowledge of time $t-1: X_{t-1},Y_{t-1},f_{t-1}$ ($X_0=Y_0=null$ and accordingly $f_0(X_0)=null$) we seek a set of samples $X$ that minimizes some function $F$, as follows.
	\begin{equation}
	X_t=\argmin_{X\subseteq\available{\samples}{t}: \mid X\mid=b}F(X, X_{t-1},Y_{t-1},f_{t-1})
	\end{equation}
\end{problem}

We propose several alternatives to the function $F$, adapting known measures for active learning to the unique characteristics of our problem. We also test different model generators $f$ in the incremental setting of active learning.

\section{Sample Collection Schedule}
\label{sec:alg}
Learning from data of the milk industry is challenging. A prime reason for that is the high cost of labeled data that require expensive physical experimentation. Therefore, we aim at making the most out of each labeled sample. To do so, we introduce in this work an \emph{active learning} approach, aiming at maximizing performance with as few labeled samples as possible~\cite{settles1995active}. With active learning, one chooses samples from which most learning can be performed.


Problem~\ref{prob} requires the generation of a sample collection schedule. In this section we propose such a schedule, partitioning it into two phases, an initialization phase in which samples are scheduled to be collected according to some pre-specified rule, and an active learning phase, in which decisions are taken based on input from previous steps. In the absence of an initial training set we propose a bootstrapping approach to support an intelligent sample selection. The initialization step does not consider a trained model or labels, and uses human meta-knowledge instead, to identify useful samples. Once a sufficient amount of samples to ensure effective learning are labeled, the learning step can begin. At one extreme, the full schedule can be determined solely using the initialization phase without actual learning. At the other extreme, initialization is performed for a single instance, after which learning comes into play.  In Section~\ref{sec:exp} we test for the best combination of the two phases. 

\ifdefined\TechReport
\begin{figure*}[htpb]
	\centering
	\includegraphics[width=0.78\linewidth]{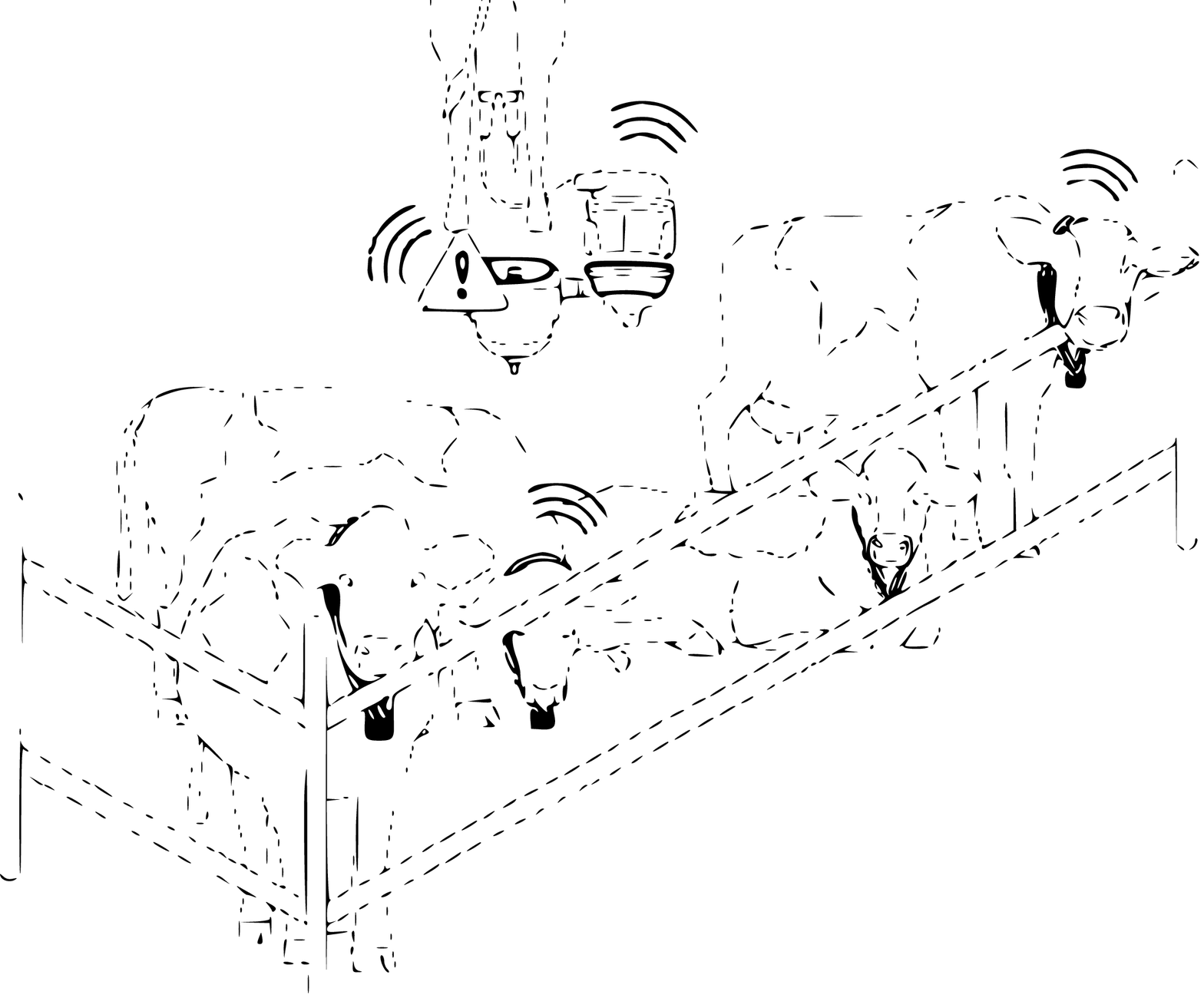}
	\caption{An Active Learning Framework}
	\label{fig:framework}
\end{figure*}
\else
\begin{figure}[htpb]
	\centering
	\includegraphics[width=.9\linewidth]{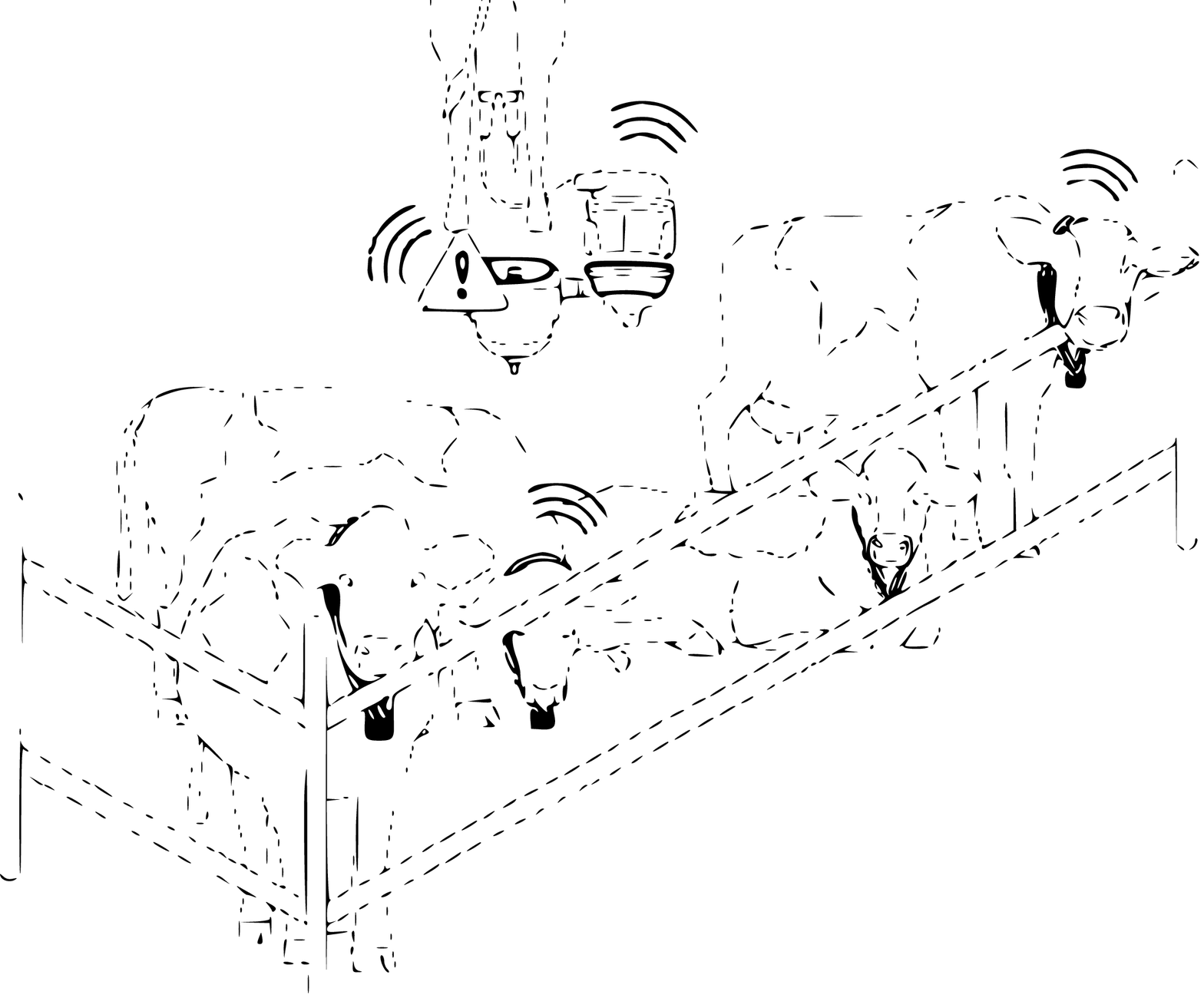}
	\caption{An Active Learning Framework}
	\label{fig:framework}
\end{figure}
\fi

Figure~\ref{fig:framework} illustrates the active learning framework we use in this work. An active learner asks queries in the form of unlabeled samples to be labeled by an oracle, which in our case is the cheese production process outcome. We iteratively construct a model based on these labels.  Using (unlabeled) preprocessed data, an initial pool of samples is selected for labeling to construct a baseline model. Then, iteratively, based on some criteria, an additional set of unlabeled samples is selected for labeling. Using these labels, the model is updated and evaluated. This process is repeated until some stopping criteria is achieved (\emph{e.g.,} the desirable performance of the model is obtained). Altogether, the key ingredient in active learning is a selection method to choose useful samples to collect and label, complying with the given sample budget.

We first introduce several methods that serve for the initialization phase (Section~\ref{sec:init}). Then, we present active learning for regression problems and focus on active learning selection methods (Section~\ref{sec:prelim}). Evaluation of the sample collection schedule is discussed in Section~\ref{sec:measures}. We conclude the section with our proposed algorithmic solution (Section~\ref{sec:alg_in}).

\subsection{Initialization Methods ($init(\cdot)$)}\label{sec:init}

In the initialization phase we choose a set of samples using some non-learning strategy. A na\"ive strategy may be to randomly choose from the pool of available samples. Next, we introduce three additional strategies that are based on some human meta-knowledge. In what follows, we present the strategies under a constraint of one sample at a time, noting that since no learning is involved between sample executions, multiple samples per iteration may be selected iteratively. Recall that in our setting, each iteration $t$ comes with its own pool of samples $\available{\samples}{t}$ and therefore each initialization step should be handled separately.

\noindent \textbf{Pareto Sampling:}\label{sec:pareto}
With domain expert assistance, we partition the feature space into two sets $\{d^{+},d^{-}\}\subseteq \{1,2,\dots d\}$ such that $d^{+}\cup d^{-}= \{1,2,\dots d\}$. $d^{+}$ and $d^{-}$ contain features that positively and negatively affect the target value, respectively.   

Using $d^{+}$ and $d^{-}$, a \emph{pareto dominance} relation between two samples $x, x^{\prime}\in \available{\samples}{t}$, is given as follows. 
$x$ {\em Pareto positively dominates} $x^{\prime}$ if the following conditions hold ($x[j]$ denotes the $j$th feature of sample $x$): (1) $\forall j\in d^{+},x[j] \geq x^{\prime}[j]$, (2) $\forall j\in d^{-},x[j] \leq x^{\prime}[j]$, and (3) $\exists j\in d^{+} \mid x[j] > x^{\prime}[j]\lor \exists j\in d^{-} \mid x[j] < x^{\prime}[j]$.

Similarly, $x$ {\em Pareto negatively dominates} $x^{\prime}$ if the following conditions hold: (1) $\forall j\in d^{+},x[j] \leq x^{\prime}[j]$, (2) $\forall j\in d^{-},x[j] \geq x^{\prime}[j]$, and (3) $\exists j\in d^{+} \mid x[j] < x^{\prime}[j]\lor \exists j\in d^{-} \mid x[j] > x^{\prime}[j]$

Pareto sampling selects samples from the pool of samples that are not (positively or negatively) Pareto dominated in $\available{\samples}{t}$.


\noindent \textbf{Distance Sampling (Di):}\label{sec:GDi}
A common sampling technique aims at distributing the samples across the whole feature space, under the assumption that samples that are close together add little information. Using geometric characteristics over the input features space, distance sampling (Di) selects a distant sample, one that is located (using some distance measure) as far as possible from already labeled samples (those in $X_{t-1}$). Formally, let $d(\cdot,\cdot)$ be a distance measure. We select a new sample $x\in \available{\samples}{t}$ such that 
\begin{equation}\label{eq:Di} 
\argmax_{x\in \available{\samples}{t}} \min\limits_{x^{\prime}\in X_{t-1}}d(x,x^{\prime})
\end{equation}
In the experiments we use Euclidean distance: $d(x,x^{\prime})=\Vert x-x^{\prime} \Vert_2$.

\noindent \textbf{Clustering sampling (Cl):}\label{sec:GDl}
Using a similar assumption as with distance sampling, we avoid using two samples that are close together. With clustering sampling, a geometric structure is used to group samples into clusters. Therefore, as a first step a clustering algorithm (\emph{e.g.,} K-means~\cite{macqueen1967some}) is utilized to generate a set of clusters over $\available{\samples}{t}$. Then, Cl selects a representative sample from each cluster and applies Di (distance sampling, see above) to choose a sample from the set of representatives. A representative may be chosen by an expert, or simply by choosing the cluster's median value (the one closest to a cluster mean). In the empirical evaluation we use the K-means algorithm with $K=20$.

\subsection{Active Learning for Regressors}
\label{sec:prelim}

Unique to our approach is the use of active learning for regression problems. Most methods in the literature address classification problems~\cite{cai2013maximizing}, assuming all incorrect labels equally err. In what follows, we now explain how to adopt existing query strategies, which were originally designed for classification problems to offer solutions to regression problems. 
While the initialization methods described above (Section~\ref{sec:init}) can continue and serve as the selection methodology, in this section we assume that we have at our disposal a set of already labeled samples $\selected{\samples}{t-1}, \selected{\labels}{t-1}$ and a respectively trained model $f_{t-1}$ when selecting samples for $\selected{\samples}{t}\subseteq \available{\samples}{t}$.

We first suggest a simple extension of Di (Eq.~\ref{eq:Di}), termed \textbf{Predictions (Pr)}, utilizing a trained model. Pr selects samples based on the distance of their predicted value, using the trained model, from other known labels. Formally, let $d(\cdot,\cdot)$ be a distance measure. We select a new sample $x\in \available{\samples}{t}$ such that 
\begin{equation}
\argmax_{x\in \available{\samples}{t}}\mathop{min}\limits_{y^{\prime}\in Y_{t-1}}d(f_{t-1}(x),y^{\prime}).
\end{equation}
\begin{sloppypar}
\noindent In the experiments we use absolute distance: $d(f_{t-1}(x),y^{\prime})=|f_{t-1}(x)-y^{\prime}|$. We divide the remaining proposed methodologies into three groups, namely, Uncertainty, QBC, and EMCM, as follows: 
\end{sloppypar}

\noindent \textbf{Uncertainty:} This selection method selects a sample at time $t$ that the current model is least certain about its label. For classification tasks, it is usually a straightforward approach, selecting a sample with the lowest margin between predicted classes, or a sample with the highest classification entropy. Such methods use the probability of a classifier to measure classification uncertainty. Switching from classification problems, where labels are discrete, to regression with labels from a continuous domain, the probability of obtaining a specific label is zero. Therefore, for a regression model, uncertainty is assessed, for example, by selecting a sample with the highest prediction variance. Note that in a regression scenario, selecting outliers may considerably change the model (even worsen performance), while for classification, it has only a minor effect on the decision boundaries~\cite{o2015evaluation}. Next, we propose three approaches to compute uncertainty for regression models:


\begin{sloppypar}
\begin{description}
	\item[Uncertainty Discretization (UDi)] discretizes the regression label, converting the label space (${\rm I\!R}$) into classes (${\rm I\!N}$, as in classification). Applying a discretization can utilize expert knowledge to determine a transformation from ${\rm I\!R}$ to ${\rm I\!N}$. Alternatively, the transformation can be done heuristically, dividing the output space into equal (weighted) bins. Then, we can transform $Y_{t-1}$ to classes and train a model $f^{\prime}_{t-1}$ over $X_{t-1}$. Once we have a trained $f^{\prime}_{t-1}$, it is used to compute the uncertainty using, {\em e.g.}, classification entropy~\cite{settles1995active}.
	
	\item[Uncertainty Clustering (UCl)] uses clustering to estimate the uncertainty of a regression model. UCl selects low quality representative samples by first applying a clustering algorithm to generate a set of clusters over $\available{\samples}{t}$ (in the experiments we used the K-means algorithm with $K=20$). Then, UCl chooses $k$ clusters having the highest variance by applying $f_{t-1}$ for each sample in a cluster. Finally, UCl uses the silhouette measure~\cite{rousseeuw1987silhouettes} to select samples with the lowest coefficient out of the chosen $k$ clusters. 
	
	\item[Uncertainty Mean Squared Error (UMSE)] is a novel approach we propose to measure regression uncertainty for a sample $x$ using a weighted mean squared error as follows:
	\begin{equation}
	UMSE(x) = \frac{\sum_{(x^{\prime},y^{\prime})\in X_{t-1}}{\frac{1}{d(x,x^{\prime})}(y^{\prime}-f_{t-1}(x^{\prime}))^{2}}}{\sum_{(x^{\prime},y^{\prime})\in X_{t-1}}{\frac{1}{d(x,x^{\prime})}}}
	\end{equation}
where $d(\cdot,\cdot)$ is a distance measure (Euclidean distance). The nominator uses inverse distances to weight the model squared errors, where larger error corresponds to higher uncertainty. The denominator is a normalization factor. Intuitively, a sample that is close to labeled samples with large squared error, has a higher uncertainty. A sample with the highest UMSE score is selected. 
\end{description}
\end{sloppypar}

\noindent \textbf{Query-by-Committee (QBC)} 
generates a committee of models from a labeled training set. A committee may include multiple trained models on different subsets of the labeled data or variations of the same model with different hyperparameter settings. A possible committee may also include models that use different algorithms (\emph{e.g.}, linear regression or random forest). 
\begin{sloppypar}
Once a committee of models is generated, the selection is based on the committee disagreement. For classification tasks, disagreement can be measured by voting entropy, selecting samples for which the entropy of the votes among the committee is the largest~\cite{dagan1995committee}. For regression tasks, the disagreement can be measured by the sum of squared error (or variance) within the predictions of the sample~\cite{burbidge2007active}. This, in turn, quantifies a finer computation of the disagreement among a committee of regressors.  
\end{sloppypar}
Formally, let $f^{1}_{t-1}, \dots, f^{C}_{t-1}$ be a set of $C$ trained models over $\selected{\samples}{t-1}$. The disagreement of the sample $x$ is $\frac{1}{C} \sum_{c=1}^{C}{(f^{c}_{t-1}(x)-\overline{y^{c}})^2}$, where $\overline{y^{c}}$ is the predictions mean of the $c$'th model.

\noindent \textbf{Expected Model Change Maximization (EMCM)} 
selects samples at time $t$ whose labels are expected to change the model the most at that time~\cite{cai2013maximizing}. A model change depends on the algorithm of choice. For example, when using linear regression, model change can be expressed as the difference between the regression line coefficients of the current model and those of the updated model. It is worth noting that such a selection is usually aimed at a faster convergence. Given that the true target value of the candidate sample is unknown, we generate a committee of models to estimate the model change. We note here that whereas QBC selects samples based on the level of disagreement within a committee, EMCM selects samples based on the level of disagreement between the committee and the actual model.

In this work we follow a method suggested by Cai {\em et al.}~\cite{cai2013maximizing}. Let $f_{t-1}$ be a model with parameters $\theta$ that minimizes some training loss and $\theta^{\prime}$ be the parameters of a model $f^{\prime}_{t-1}$ that minimizes the loss by adding the sample $(x,y)$ to $\selected{\samples}{t-1}$. Then, the model change is given by $\Vert \theta-\theta^{\prime} \Vert_2^2$, measuring the difference between the model parameters with and without the sample $x$. To set the proper set of parameters $\theta^{\prime}$, we use stochastic gradient descent~\cite{bottou2010large}, according to which, we update the model parameters $\theta$ following a negative trend of the training loss. Specifically, if $f_{t-1}$ is a linear regression model, $\theta$ is the coefficients and its loss is given by MSE (see Section~\ref{sec:measures}). The MSE of the sample $(x,y)$ is $(\theta^T x-y)^2$ and using SGD we obtain $\theta^{\prime}=\theta-2\mu x(\theta^T x-y)$ where $\mu$ is some learning rate. Accordingly, The model change is $\Vert 2\mu x(\theta^T x-y)\Vert _2^2$. We estimate the true model change using a committee of $C$ models, as follows: $\frac{1}{C}\sum_{c=1}^{C} \Vert 2\mu x(\theta^T x-f^{c}_{t-1}(x) ) \Vert _2^2 $ and select that sample that maximizes it.

\subsection{Active Learning Evaluation}\label{sec:measures}

Accompanied by the variety of active learning methods, there is a need to evaluate the learning process and compare the performance of the various methods. Shen \emph{et al.} define a {\em useful sample} using three qualitative criteria, namely, informativeness, representativeness, and diversity~\cite{shen2004multi,wu2018pool}. \emph{Informativeness} dictates that selected samples should contain rich information, to allow the model to significantly benefit from labeling them. \emph{Representativeness} suggests selecting samples that represent a large portion of the input data~\cite{huang2010active} and \emph{diversity} states that samples should scatter across the full input space. Overall, representativeness and diversity ensure coverage of a large portion of the input space and good-sized trained data call for informativeness to enrich the model and certify its validity.

\ifdefined\TechReport
\begin{table*}[htpb]
	\caption{Comparison between active learning methods. For each method we specify whether it uses a prior trained model ($f_{t-1}$), an existing training set ($X_{t-1}$) and the labels of an existing training set ($Y_{t-1}$). We also state whether the method requires an expert knowledge, complies with one of three qualitative criteria~\cite{shen2004multi}, and subjectively assess its computation time.}\label{tab:methods}
	\begin{tabular}{|l l|c|c|c|c|c|c|c|c|c|}
		\hline
		~ & ~ & Pareto & Di & Cl & Pr & UDi & UCl & UMSE & QBC & EMCM \\ \hline
		~ &Uses $f_{t-1}$ & \xmark & \xmark & \xmark & \cmark & \xmark & \cmark & \cmark & \xmark & \cmark \\ 
		Usage &Uses $X_{t-1}$ & \xmark & \cmark & \cmark & \xmark & \cmark & \xmark & \cmark & \cmark & \cmark  \\ 
		~ &	Uses $Y_{t-1}$ & \xmark & \xmark & \xmark & \cmark & \cmark & \xmark & \cmark & \cmark & \cmark  \\  
		~ &Uses Expert Knowledge & \cmark & \mmark & \mmark & \mmark & \cmark & \mmark & \mmark & \xmark & \xmark  \\ \hline
		~ &Diversity & \cmark & \cmark & \cmark & \cmark & \xmark & \xmark & \xmark & \xmark & \xmark  \\ 
		Evaluation &Representativeness & \xmark & \xmark & \cmark & \xmark & \xmark & \mmark & \xmark & \xmark & \xmark \\ 
		~ &Informativeness & \xmark & \xmark & \xmark & \cmark & \cmark & \cmark & \cmark & \cmark & \cmark \\ \hline
		Performance &Computation Time & \emph{F} & \emph{F} & \emph{F} & \emph{F} & \emph{M} & \emph{M} & \emph{F} & \emph{S} & \emph{S}  \\ 
		\hline
	\end{tabular}
\end{table*}

\else
\begin{table}[htpb]
	\caption{Comparison between active learning methods. For each method we specify whether it uses a prior trained model ($f_{t-1}$), an existing training set ($X_{t-1}$) and the labels of an existing training set ($Y_{t-1}$), whether the method requires an expert knowledge, complies with one of three qualitative criteria~\cite{shen2004multi}, and subjectively assess its computation time.}\label{tab:methods}
	\scalebox{0.75}{\begin{tabular}{|l|c|c|c|c|c|c|c|c|c|}
			\hline
			~ & Pareto & Di & Cl & Pr & UDi & UCl & UMSE & QBC & EMCM \\ \hline
			Uses $f_{t-1}$ & \xmark & \xmark & \xmark & \cmark & \xmark & \cmark & \cmark & \xmark & \cmark \\ 
			Uses $X_{t-1}$ & \xmark & \cmark & \cmark & \xmark & \cmark & \xmark & \cmark & \cmark & \cmark  \\ 
			Uses $Y_{t-1}$ & \xmark & \xmark & \xmark & \cmark & \cmark & \xmark & \cmark & \cmark & \cmark  \\  
			Uses An Expert & \cmark & \mmark & \mmark & \mmark & \cmark & \mmark & \mmark & \xmark & \xmark  \\ \hline
			Diversity & \cmark & \cmark & \cmark & \cmark & \xmark & \xmark & \xmark & \xmark & \xmark  \\ 
			Representativeness & \xmark & \xmark & \cmark & \xmark & \xmark & \mmark & \xmark & \xmark & \xmark \\ 
			Informativeness & \xmark & \xmark & \xmark & \cmark & \cmark & \cmark & \cmark & \cmark & \cmark \\ \hline
			Computation Time & \emph{F} & \emph{F} & \emph{F} & \emph{F} & \emph{M} & \emph{M} & \emph{F} & \emph{S} & \emph{S}  \\ 
			\hline
	\end{tabular}}
\end{table}
\fi

Table~\ref{tab:methods} provides a qualitative comparison between the different methodologies. For each method we specify whether it utilizes $f_{t-1}$, $X_{t-1}$, $Y_{t-1}$, whether it uses expert knowledge, with which qualitative criteria~\cite{shen2004multi} it complies, and an approximate computation time. with \emph{S} corresponding to slow, \emph{M} to moderate, and \emph{F} to fast.

Pareto requires no prior knowledge of the domain as determined by $f_{t-1}$, $X_{t-1}$, $Y_{t-1}$ and only heuristically defines a ranking among samples. UMSE and EMCM use all possible resources when selecting samples. Most methods can benefit from expert knowledge, except QBC and EMCM, which rely on multiple trained models. All selection methods comply with informativeness as they utilize $f_{t-1}$, $Y_{t-1}$ or both. Cl adheres to the representativeness as it chooses a representative among each cluster. Initialization methods as well as Pr choose diverse samples using the space of samples as a guiding principle in selecting samples. Finally, considering weekly cycles of learning (as in our case study) the importance of computation time is negligible. Nevertheless, we note that while initialization methods are comparatively fast, committee-based methods (QBC and EMCM) are quite slow as they require training multiple models.

\subsubsection{Qualitative Evaluation Measures}

The standard evaluation measure for regressors is mean squared error (MSE)~\cite{settles1995active}, or one of its derivatives, \emph{e.g.,} RMSE. Let $Y$ and $\hat{Y}$ be a set of outcomes and a set of predicted outcomes, respectively. MSE is given by:
\begin{equation}\label{sec:mse}
MSE(Y, \hat{Y}) = \frac{1}{n}\sum_{i=1}^{n} (Y_{i} - \hat{Y}_{i})^2
\end{equation}

MSE evaluates a set of samples jointly, regardless of the evaluation timing of samples. To better understand the quality of active learning as a process of sample selection, we suggest next measures that are more suited to quantify temporal cumulative quality by assessing aggregated performance, convergence, and smoothness.      

We use Area Under Curve (AUC) to measure the MSE behavior over the learning course and evaluate the performance of active learning methods over time~\cite{baram2004online}. AUC is used for classification problems but can also be applied in regression tasks. AUC is a measure of error, so the lower the AUC, the better the method. Yet, since MSE is not bounded for regression, the AUC is extremely exposed to outliers. To overcome this deficiency, we propose the use of the log of AUC (logAUC). Let $MSE_{t}(Y, \hat{Y})$ be the MSE of $Y$ and $\hat{Y}$ at the $t$-th timestamp. AUC and logAUC are defined as follows: 
\begin{equation}\label{sec:auc}
	\begin{split}
		AUC(Y, \hat{Y}) & = \sum_{t=1}^{T} MSE_{t}(Y, \hat{Y})\\
		logAUC(Y, \hat{Y}) & = \sum_{t=1}^{T} log(MSE_{t}(Y, \hat{Y}) + 1)
	\end{split}
\end{equation} 

To measure smoothness, we adapt the Absolute Second Difference (ASD) measure~\cite{lappi2006smooth}, which uses second differences to capture the inflexion of the curve~\cite{barnett1985criteria}. ASD uses $MSE_{t}(Y, \hat{Y})$ as follows: 

\begin{equation}\label{sec:smooth1}
	\begin{aligned}
		ASD(Y, \hat{Y}) = &\frac{1}{T-1} \sum_{t=2}^{T-1} |MSE_{t+1}(Y, \hat{Y})\\
		& - 2MSE_{t}(Y, \hat{Y}) + MSE_{t-1}(Y, \hat{Y})|
	\end{aligned}
\end{equation} 

Early stages of learning may demonstrate unstable behavior as they only rely on a small set of samples. To compensate for such instability, we also use a weighted version of ASD, as follows:
\begin{equation}\label{sec:smooth2}
\begin{aligned}
WASD(Y, \hat{Y}) = &\frac{2}{(T-1)(T-2)} \sum_{t=2}^{T-1} t\cdot |MSE_{t+1}(Y, \hat{Y})\\
                   & - 2MSE_{t}(Y, \hat{Y}) + MSE_{t-1}(Y, \hat{Y})|
\end{aligned}
\end{equation} 

Finally, to measure convergence, we use a First to Converge (FTC) measure (inspired by the definition of a sequence limit), quantifying the extent of which the learning converges to a small difference ($\varepsilon>0$) between consecutive iterations. Using $MSE_{t}(Y, \hat{Y})$, let $\varepsilon$ be a small positive number.  FTC is defined as follows:
\begin{equation}\label{sec:conv}
	\begin{aligned}
		FTC(Y, \hat{Y}) &= \mathop{argmin}\limits_{1\leq t\leq n} \{t: |MSE_{i}(Y, \hat{Y})\\
		& - MSE_{i-1}(Y, \hat{Y})|<\varepsilon, \forall t<i\leq n\}
	\end{aligned}
\end{equation} We note that in the experiments we use $\varepsilon=0.01$.

\subsection{Algorithm}\label{sec:alg_in}

Algorithm~\ref{alg:main} details our methodology, adopting active learning, of training using data that is gathered from IoT sensors. According to the setting of our case study (Section~\ref{sec:dairy}), our aim is to select samples for learning on a weekly basis. The input of the algorithm is a sequence of available sample sets over an experiment horizon $\available{\samples}{1}, \available{\samples}{2}, \dots, \available{\samples}{T}$ and an hyperparameter $K\leq T$ representing the number of initialization steps. First we apply one of the initialization methodologies (see Section~\ref{sec:init}) for $K$ steps (lines 4-7), based on which we train an initial model (line 8). 
\ifdefined\TechReport
We analyze the implementation of $init(\cdot)$ in Section~\ref{sec:initexp}, choosing a methodology as well as the size of $K$.
\else
We analyze the implementation of $init(\cdot)$ in a technical report~\cite{tech}.
\fi

\begin{algorithm}[h!]
	\begin{algorithmic}[1]
		\State \textbf{Input:} $\available{\samples}{1}, \available{\samples}{2}, \dots, \available{\samples}{T}$, $K<T$ 
		\State \textbf{Output:} A trained model $f$ 
		\State $X_0=Y_0=null$
		\For{$t=1$ to $K$}\Comment Initialization.
		\State $\selected{\samples}{t} \leftarrow select(\available{\samples}{t})$ 
		\State $\selected{\labels}{t} \leftarrow$ Obtain labels for $\selected{\samples}{t}$
		\State $\selected{\samples}{t} =\selected{\samples}{t}\cup \selected{\samples}{t-1};$ $\selected{\labels}{t} = \selected{\labels}{t}\cup \selected{\labels}{t-1}$
		\EndFor
		\State Train a model $f_{K}$ using $\selected{\samples}{K}, \selected{\labels}{K}$ \Comment Constructing an initial model.
		\For{$t=K+1$ to $T$}\Comment Active learning. 
		\State $\selected{\samples}{t} \leftarrow select(\available{\samples}{t})$ 
		\State $\selected{\labels}{t} \leftarrow$ Obtain labels for $\selected{\samples}{t}$
		\State $\selected{\samples}{t} =\selected{\samples}{t}\cup \selected{\samples}{t-1};$ $\selected{\labels}{t} = \selected{\labels}{t}\cup \selected{\labels}{t-1}$
		\State Train a model $f_{t}$ using $\selected{\samples}{t}, \selected{\labels}{t}$
		\EndFor
		\State \textbf{Return:}  $f = f_{T}$ 
	\end{algorithmic}
	\caption{Data Collection Schedule Algorithm}\label{alg:main}
\end{algorithm}

Once an initial model is obtained, we utilize advanced active learning methodologies, which require an already trained model (repeating lines 9-13 for $T-K$ steps). Specifically, each round begins with selecting a set of samples for labeling $\available{\samples}{t}\subseteq\selected{\samples}{t}$ (lines 10-11) using a selection methodology $select(\cdot)$ (out of the ones presented in Section~\ref{sec:prelim}). Then, the new labeled samples are added to the accumulated labeled sample set $\selected{\samples}{t-1}$ (line 12), which is then used to train a new model (line 13). Finally, after a stopping criterion is reached, we return the current model as the output (line 14).  

\section{Empirical Evaluation}
\label{sec:exp}
We conducted a thorough empirical evaluation and report on its results herein. We start with a description of the experimental setup (Section~\ref{sec:setup}), followed by report of results (Section~\ref{sec:results}).

\subsection{Experimental Setup}\label{sec:setup}
We first describe our experimental design with an overarching goal of understanding the impact of milk features on the production of cottage cheese. We introduce the dataset based on our case-study (Section~\ref{sec:useCase}) and detail the experiments methodology. We further provide the code and a sample of the data in a repository~\cite{gitURL}.

\subsubsection{Dataset}
The dataset is composed of two daily reports from the milk farm, one characterizing the cows and the other the milk itself. Each sample represents a cow at a date, such that multiple cows data it available at a single date and different instances of the same cow are available at different dates. 

Features of cows include age, number of lactations, diseases, and an activity report including number of strides and recumbence.
Milk features include quantity, milking time, milking rate, and chemical content including protein, fat, lactose, somatic cells, and urea. 
Most features are numeric with a few exceptions of binary values. Missing values were imputed using historical data. 

The proposed algorithm was implemented in a real-world experiment, jointly with a large dairy in Israel. Twenty-four Israeli Holstein Herds at different sizes varying from $80$ to $,1000$ individual cows participate in the study. The total daily production of the herds is approximately $400,000$ Kg. The data of the individual cows comprising each herd is aggregated to herd data presentation. Once a week, our data scheduler (Algorithm~\ref{alg:main}) selects herds (samples) with production amounting to $\sim 100,000$ Kg daily. This milk is collected by the dairy and channeled to produce cottage cheese. Commercial Protein efficiency and Cheese Yield are the labels (targets) in this setting. Due to third-party confidentiality restrictions on dairy performance measures, we used Curd Firmness (CF, $10.55 \pm 7.89$), as the regression label ($Y$), calculated at Afimilk laboratory (see Section~\ref{sec:dairy}), and used individual cows as samples for evaluating the algorithm various settings. 

\subsubsection{Methodology}

Afimilk weekly reports, containing samples of 500 cows ($\forall t\mid\available{\samples}{t}\mid=500$), were used to evaluate Algorithm~\ref{alg:main}. The labeling budget for each iteration is 8 cows ($\forall t\mid\selected{\samples}{t}\mid=8$). We experimented with $T=100$ iterations, from which we allocated a maximum of 25 iterations for initialization (we validated the span $K\in\{5,10,15,20,25\}$). In addition, for validation purposes, we use an \textbf{unseen} test set to test the performance of the different methods. 
\ifdefined\TechReport
We analyze the following aspects of Algorithm~\ref{alg:main}: 
\begin{sloppypar}
\noindent\textbf{Regression method (Section~\ref{sec:regexp}),} examining linear~\cite{seber2012linear}, Random Forest~\cite{liaw2002classification}, XGBoost~\cite{chen2016xgboost}, and Multi-layer Perceptron neural network~\cite{glorot2010understanding} regressors.\\
\noindent\textbf{Initialization method (Section~\ref{sec:initexp}),} optimizing the number of  initialization steps and the methods from Section~\ref{sec:init}.\\
\noindent\textbf{Selection method (Section~\ref{sec:selectexp}),} testing the methods described in Section~\ref{sec:prelim}.\\
\noindent\textbf{Active learning quality (Section~\ref{sec:qualityexp}),} using the measures described in Section~\ref{sec:measures}.\\
\noindent\textbf{Features importance analysis (Section~\ref{sec:featuresexp}),} using SHAP~\cite{shap}.
\end{sloppypar}
\else
Due to space considerations, some analyses are not provided here. Specifically, a \textbf{regression method} analysis (examining linear~\cite{seber2012linear}, Random Forest~\cite{liaw2002classification}, XGBoost~\cite{chen2016xgboost}, and Multi-layer Perceptron neural network~\cite{glorot2010understanding} regressors), an \textbf{initialization method} analysis (optimizing the number of  initialization steps and the methods from Section~\ref{sec:init}), and a \textbf{selection method} analysis (testing the methods described in Section~\ref{sec:prelim}) are given in a technical report~\cite{tech}. The main results regarding the \textbf{active learning quality} (using the measures described in Section~\ref{sec:measures}) and \textbf{features importance analysis} (using SHAP~\cite{shap}), are given in Section~\ref{sec:results}.
\fi

We tested 
both a bootstrap committee (QBC$_{boot}$ and EMCM$_{boot}$) and a models committee (QBC$_{model}$ and EMCM$_{model}$). For the latter, we use Ridge, Lasso, Linear regression, Random Forest, Gradient Boosting Machine, and KNN regressor models.  

For linear regression, random forest, and multi-layer perceptron implementation we use scikit-learn\footnote{\url{https://scikit-learn.org/stable/}} and for XGBoost we use a Python implementation\footnote{\url{https://xgboost.readthedocs.io/en/latest/python/python_intro.html}} with default parameter settings. To allow a fair comparison, we fixed the initial seed. To avoid a ``cherry picking'' effect, we ran the  na\"ive strategy of random selection 15 times and report on the average performance.

Following Baram {\em et al.}~\cite{baram2004online}, we use the logAUC to compare the performance of learning among the methods. Recall that the reported results are based on an unseen test set. In order to fairly compare and select among the different methods, we checked statistical significance with a paired two-tailed t-test to compare the $\log(MSE+1)$ values along the experiment horizon (with $p$-$value<.05$). When comparing multiple methods, we also add a Bonferroni correction. We have computed all significance tests comparing among methods and report on the most dominant ones throughout this section. Table~\ref{tab:quality} presents statistical significance differences of performance in terms of logAUC over random sampling.

\subsection{Results}\label{sec:results}

Active learning is the methodology of choice for this work, due to the scarce availability of labeled data. The goal of our research is to predict milk-cheese quality and understand the effect of raw material on production outcome. 
\ifdefined\TechReport
In sections~\ref{sec:regexp}-\ref{sec:selectexp} we analyze the parameters of the best active learning method to achieve this goal, testing type of regression (Section~\ref{sec:regexp}), initialization method (Section~\ref{sec:initexp}) and selection method (Section~\ref{sec:selectexp}). Then, in Section~\ref{sec:qualityexp} we analyze the quality of learning and finally, in Section~\ref{sec:featuresexp}, we examine the effect of raw material, which are encoded as features (see Section 3.1), on the quality.
\else
After a thorough analysis, given in a technical report~\cite{tech}, we perform 15 initialization iterations of Distance Sampling (Di-15, Section~\ref{sec:init}) after which a random forest regression is set as the regressor of choice. The technical report also elaborates on choosing a selection method. In Section~\ref{sec:qualityexp} we analyze the quality of learning and in Section~\ref{sec:featuresexp}, we examine the effect of raw material, encoded as features, on the quality. 
\fi

\ifdefined\TechReport
\subsubsection{Choosing a Regressor}\label{sec:regexp}

We begin by comparing the performance of different regressors. \add{The Multi-layer Perceptron model yielded poor results when compared to the other tested regressors with an average of $401$ logAUC over the examined settings. In comparison, linear regression (see Figure~\ref{fig:reg}), has yielded an average logAUC of $290$. Thus, we exclude the Multi-layer Perceptron from the following analysis. The inferior results may suggest that when the dataset size is limited, as in the case of active learning, the dominating abilities of deep learning are neutralized.}

A comparison between linear regression, Random Forest regression, and XGBoost regression based on logAUC is given in Figure~\ref{fig:reg}. We tested each regressor with a varying number of initialization steps and all initialization methods (Section~\ref{sec:init}), as well as random sampling and a setting with no initial step (total of 252 executions per regressor). Each combination of an initialization method, number of initialization steps, and a selection method is represented as a single dot in the figure.

\begin{figure}[t]
	\centering
	\includegraphics[width=.5\textwidth]{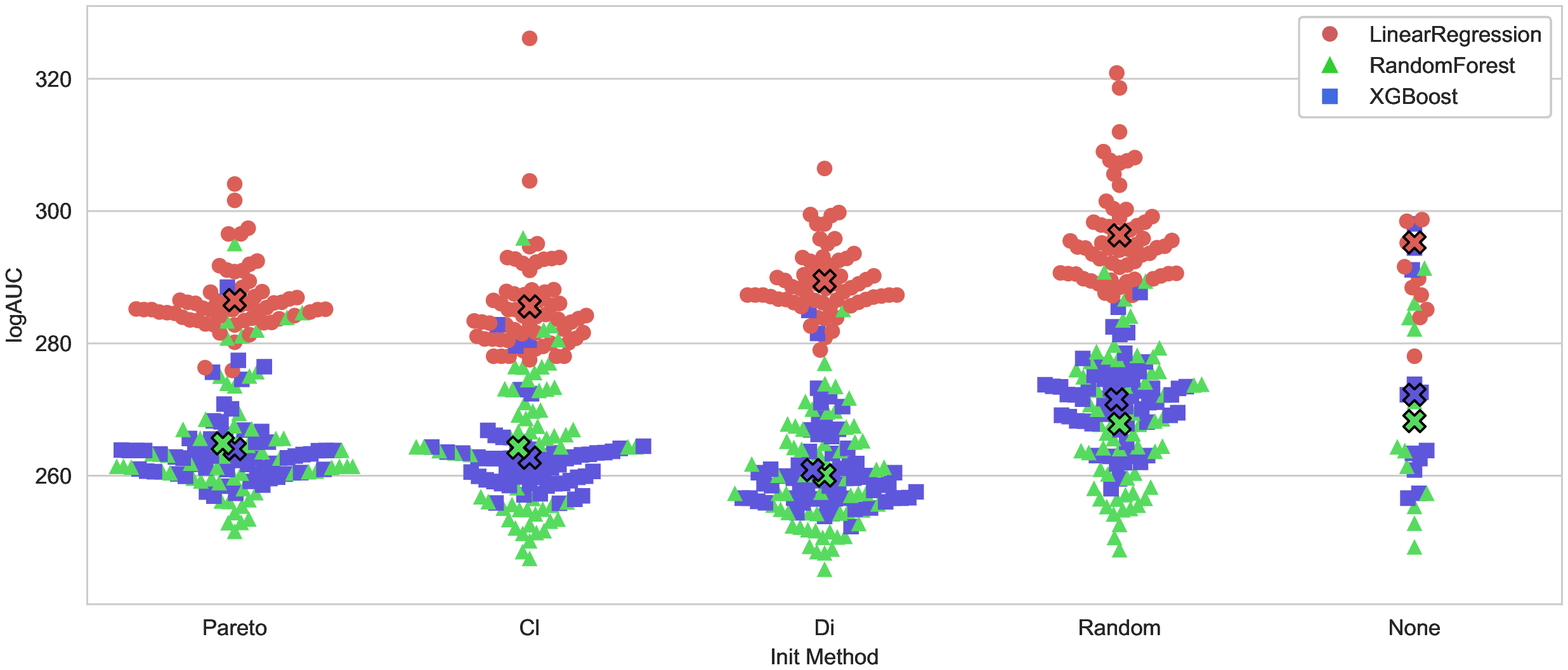}
	\caption{Comparison between the different regressors, partitioned based on initialization methods (x-axis) and evaluated using logAUC score (y-axis). The ``X'' denotes an average value of a regressor per initialization method.}
	\label{fig:reg}
\end{figure}

Figure~\ref{fig:reg} clearly shows that the random forest and XGBoost regressors outperform linear regression (results are statistically significant). Overall, the performance of random forest and XGboost over Cl, Di and Pareto is quite similar (results are not statistically significant). Yet, when looking at the performance over random sampling and no initialization, random forest achieves a statistically significant improvement in performance over XGboost. Additionally, when looking at the best results (lower part of the plot), random forest frequently achieves lower error rates. Lower error rates are obtained by the dominate selection methods (to be evaluated in Section~\ref{sec:selectexp}) involving EMCM and QBC (Section~\ref{sec:prelim}). Henceforth, we shall use random forest regression as the regressor of choice.

\subsubsection{Choosing an Initialization Method}\label{sec:initexp}

We examine the different initialization methods described in Section~\ref{sec:init}, namely, Pareto, Distances (Di), and Clustering (Cl). We compare the three methods to random sampling (Random) and no initialization (None). We also analyze the length of the initialization phase in \{5,10,15,20,25\}. Table~\ref{tab:init} compares the MSE of the different initialization methods over time as well as the logAUC (based on which we select the method, see Section~\ref{sec:setup}). The reported results are averaged over the top performing selection methods (to be evaluated in Section~\ref{sec:selectexp}).

\begin{table*}[htpb]
	\centering
	\caption{Comparison of initialization methods (Section~\ref{sec:init}) in terms of MSE over time and logAUC. Statistical significance logAUC performance over random (based on the same initialization steps) is marked with an asterisk}
	\label{tab:init}
	\scalebox{1}{\begin{tabular}{|l|c|cccccccc|c|} 
		\hline
		$init(\cdot)$          & \multicolumn{1}{l|}{\# steps} & 5               & 10              & 15              & 20              & 25              & 50             & 75             & 100            & \multicolumn{1}{l|}{logAUC}  \\ 
		\hline
		\multirow{5}{*}{Cl}     & 5                             & \textbf{22.08}  & 19.41          & 22.25           & \textbf{12.77}  & 12.04           & 10.43          & 9.55           & 9.30           & 253.40                       \\
		& 10                            & 22.08           & 17.98           & 22.54           & 13.73           & 12.23           & 10.37          & 9.74           & 9.18           & 254.60                       \\
		& 15                            & 22.08           & 17.98           & \textbf{14.69}  & 13.49           & 11.87           & 10.32          & 9.50           & 9.20           & 251.82$^{*}$                       \\
		& 20                            & 22.08           & 17.98           & 14.69           & 15.54           & 12.49           & 10.65          & 10.00          & 9.83           & 256.00                       \\
		& 25                            & 22.08           & 17.98           & 14.69           & 15.54           & 14.70           & 10.61          & 9.88           & 9.75           & 256.38                       \\ 
		\hline
		\multirow{5}{*}{Di}     & 5                             & 26.46           & \textbf{17.76}  & 19.68           & 14.46           & 12.96           & 10.33          & 9.51           & 9.25           & 253.78                       \\
		& 10                            & 26.46           & 19.01           & 18.59           & 14.41           & 12.90           & 10.04          & 9.40           & \textbf{9.00}  & 252.75$^{*}$                        \\
		& 15                            & 26.46           & 19.01           & 19.36           & 13.70           & 11.97           & 9.68           & \textbf{9.22}  & 9.11           & \textbf{250.55}$^{*}$               \\
		& 20                            & 26.46           & 19.01           & 19.36           & 17.25           & \textbf{11.45}  & \textbf{9.59}  & 9.43           & 9.30           & 252.54$^{*}$                        \\
		& 25                            & 26.46           & 19.01           & 19.36           & 17.25           & 14.59           & 9.62           & 9.32           & 9.47           & 253.61$^{*}$                        \\ 
		\hline
		\multirow{5}{*}{Pareto} & 5                             & 22.64           & 20.04           & 21.89           & 14.32           & 13.84           & 10.35          & 9.89           & 9.69           & 257.39                       \\
		& 10                            & 22.64           & 22.83           & 17.64           & 14.17           & 13.78           & 10.38          & 9.83           & 9.32           & 256.39                       \\
		& 15                            & 22.64           & 22.83           & 32.12           & 14.75           & 13.70           & 10.57          & 9.90           & 9.54           & 259.74                       \\
		& 20                            & 22.64           & 22.83           & 32.12           & 16.55           & 12.31           & 10.10          & 9.87           & 9.55           & 257.75                       \\
		& 25                            & 22.64           & 22.83           & 32.12           & 16.55           & 15.68           & 10.45          & 10.01          & 9.60           & 260.33                       \\ 
		\hline
		\multirow{5}{*}{Random} & 5                             & 22.56           & 21.97           & 19.78           & 14.89           & 12.81           & 10.49          & 10.00          & 9.46           & 255.42                       \\
		& 10                            & 22.56           & 21.35           & 25.36           & 16.33           & 13.19           & 10.50          & 9.63           & 9.15           & 257.54                       \\
		& 15                            & 22.56           & 21.35           & 20.91           & 16.42           & 13.35           & 10.89          & 9.69           & 9.39           & 259.36                       \\
		& 20                            & 22.56           & 21.35           & 20.91           & 17.69           & 13.87           & 10.69          & 9.68           & 9.21           & 258.63                       \\
		& 25                            & 22.56           & 21.35           & 20.91           & 17.69           & 16.43           & 10.83          & 9.81           & 9.47           & 260.37                       \\ 
		\hline
		None                    & 0                             & 25.33           & 25.83           & 19.27           & 14.67           & 13.00           & 9.95           & 9.61           & 9.13           & 255.23                       \\
		\hline
	\end{tabular}}
\end{table*}

 Table~\ref{tab:init} shows all three initialization methods to perform better than random sampling and no initialization. Overall, Di-15 (\emph{i.e.,} applying Distances with 15 steps of initialization) yielded the lowest average logAUC,\footnote{this result is not statistically significant compared to all methods using Bonferroni correction. Yet, it is statistically significant compared to random and the second best method, Cl-15.} and thus, henceforth we use it as our initialization method of choice. Zooming in on the performance of Di and Cl until the $15$th iteration, Cl (and specifically Cl-15) achieves a better MSE. The behavior of Di and Cl is quite similar until the $40$th iteration, after which Di becomes consistently better. It is worth noting that even though Pareto has a slow start, towards the end it achieves comparable MSE to Di.
 
 In Figure~\ref{fig:init} we compare the performance of the top performing version of each method with respect to the number of initialization steps over time. The first 10 iterations are noisy regardless of the method. This may be explained by the vulnerability to outliers having a small training set, which characterizes the initial phase. Starting at the $15$th iteration, the initialization methods stabilize and consistently decrease. Di-15 seem to converge quickly (around the $60$th iteration) and outperform all other methods from that point and on.   

 \begin{figure*}[htpb]
	\centering
	\includegraphics[width=0.35\linewidth, height=0.305\linewidth]{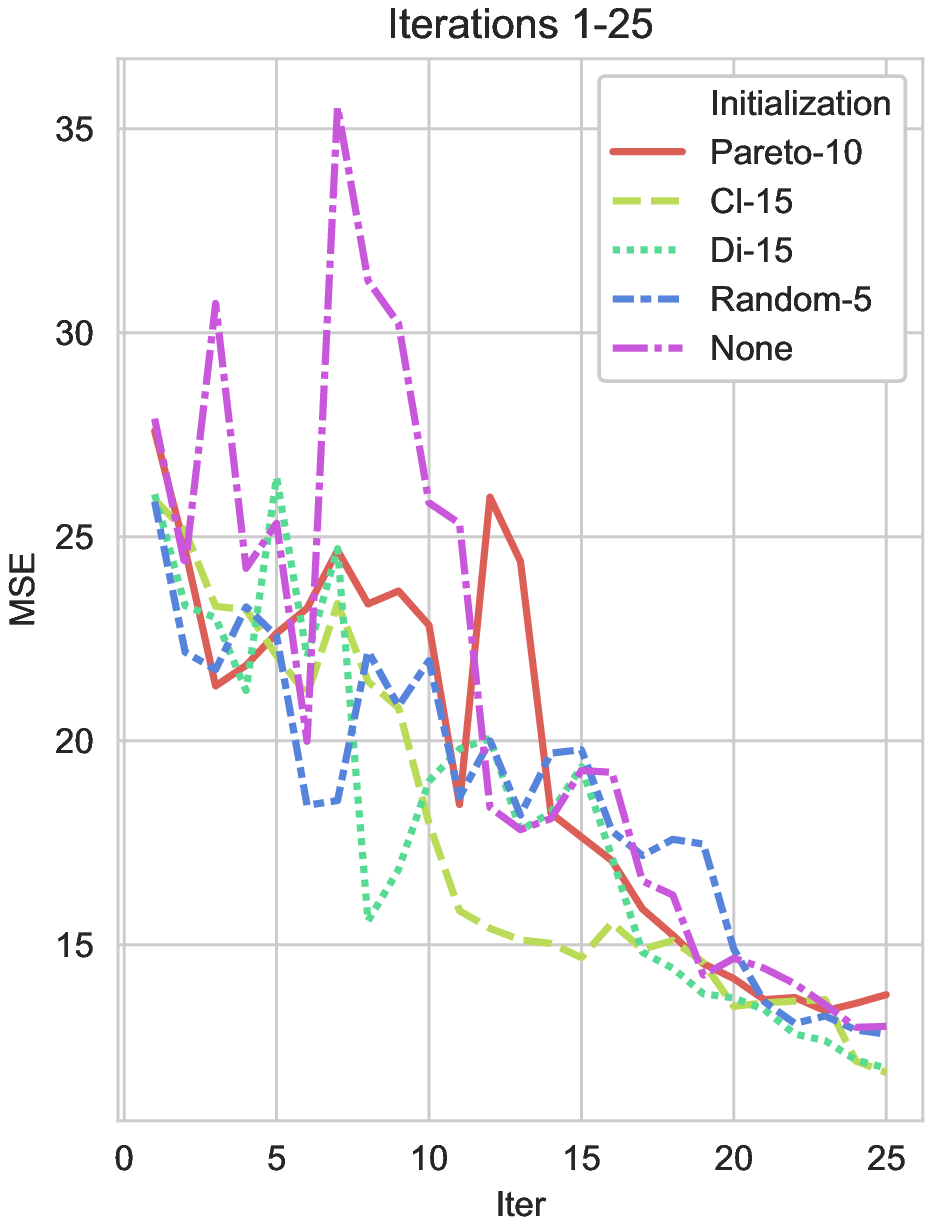}
	\includegraphics[width=0.64\linewidth]{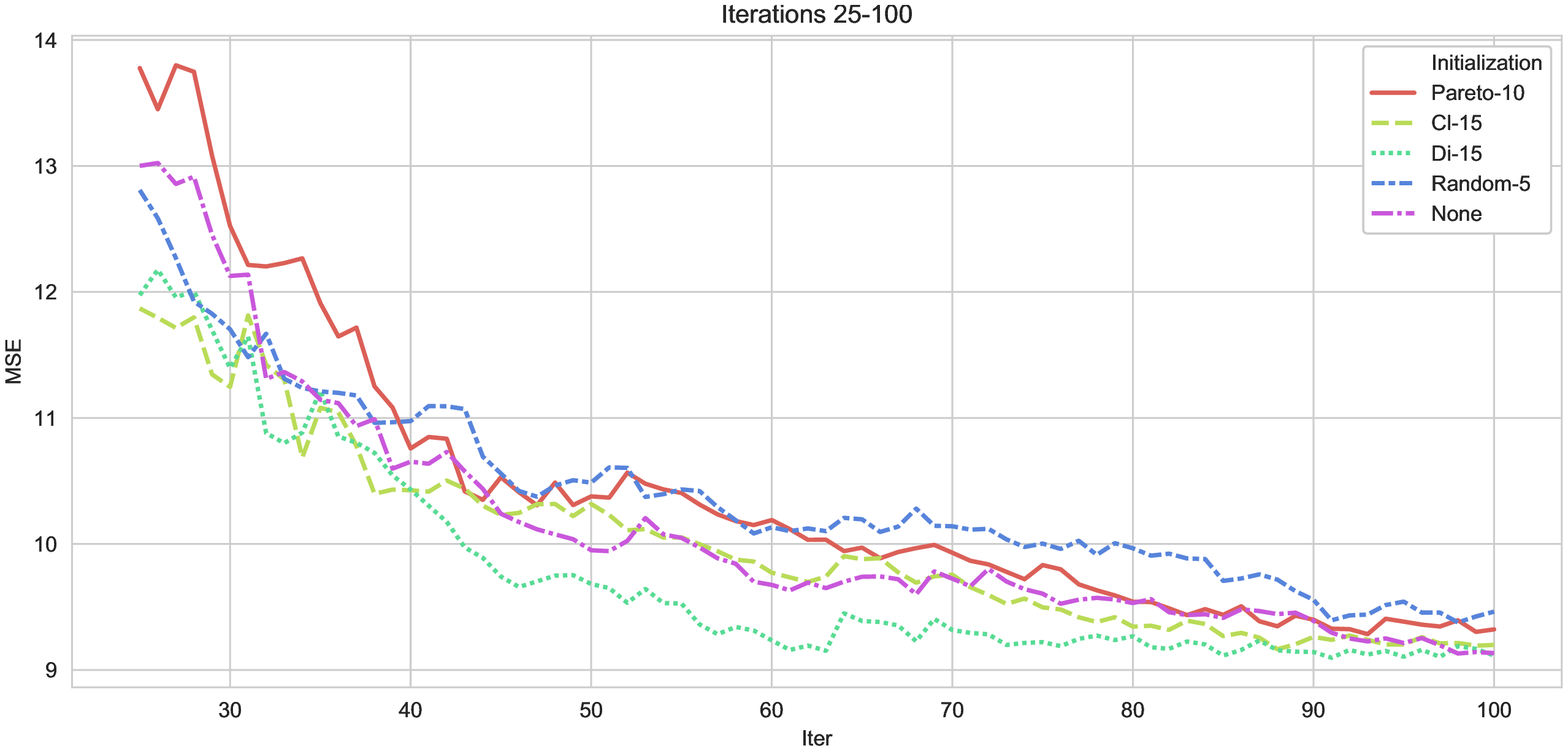}
	\caption{Comparison of initialization methods (Section~\ref{sec:init}) over time, where we present the top performing version of each method with respect to the number of initialization steps. The x-axis represents the iteration number and the y-axis the MSE score. The initialization (left) is presented along side an average performance of the different selection methods over the remaining iterations (right).}
	\label{fig:init}
\end{figure*}

\subsubsection{Choosing a Selection Method}\label{sec:selectexp}

Next, we analyze the active learning selection methods. Figure~\ref{fig:select} illustrates the MSE of the examined selection methods, from which we show the best performing version based on their logAUC (see Figure~\ref{fig:selectAUC}). Specifically, the illustrated methods are random sampling, Expected Model Change Maximization (EMCM$_{model}$), Predictions (Pr), Uncertainty Mean Squared Error (UMSE), and Query-by-Committee (QBC$_{boot}$).

\begin{figure}[htpb]
	\centering
	\includegraphics[width=\linewidth]{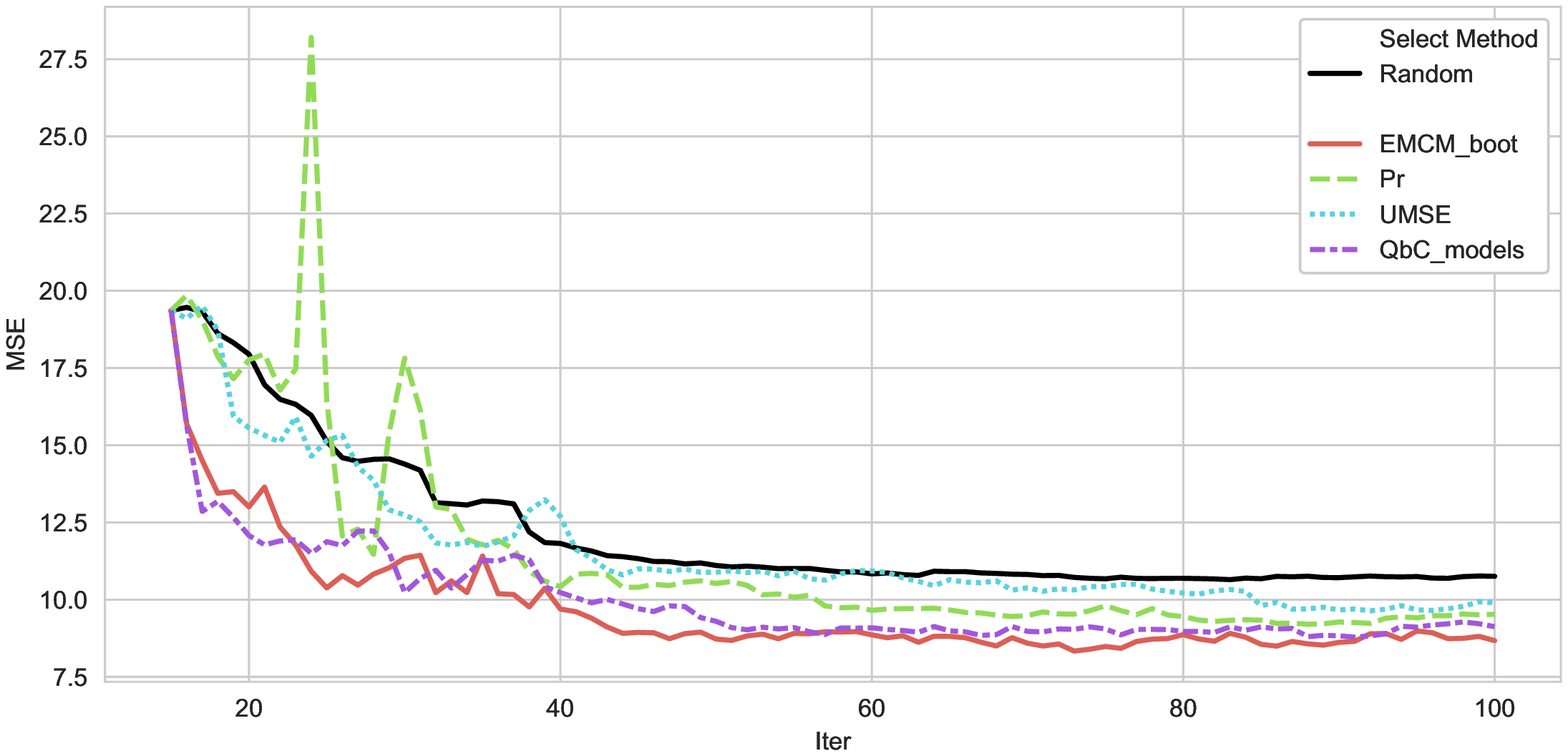}
	\caption{Comparison between selection methods (Section~\ref{sec:prelim}) over time using a random forest regressor and 15 initialization steps using Di. For each method we show the version providing the lowest logAUC (see Figure~\ref{fig:selectAUC}). The x-axis represents the iteration number (starting after the initialization) and the y-axis the MSE score.}
	\label{fig:select}
\end{figure}

As illustrated in Figure~\ref{fig:select}, QBC and EMCM methods provide a notable improvement in MSE after the initialization phase and outperform the other methods in terms of MSE throughout. This emphasizes the value of committees in selecting useful samples. UMSE performs better than random and consistently decreases over time; yet, unable to compete with QBC and EMCM. This may suggest that uncertainty methods are harder to adopt for regression problems as they typically quantify the certainty of a discrete value. Finally, we note that Pr starts out noisy and stabilizes around the $30$th iteration, outperforming both random and UMSE.   

\begin{figure}[htpb]
	\centering
	\includegraphics[width=\linewidth]{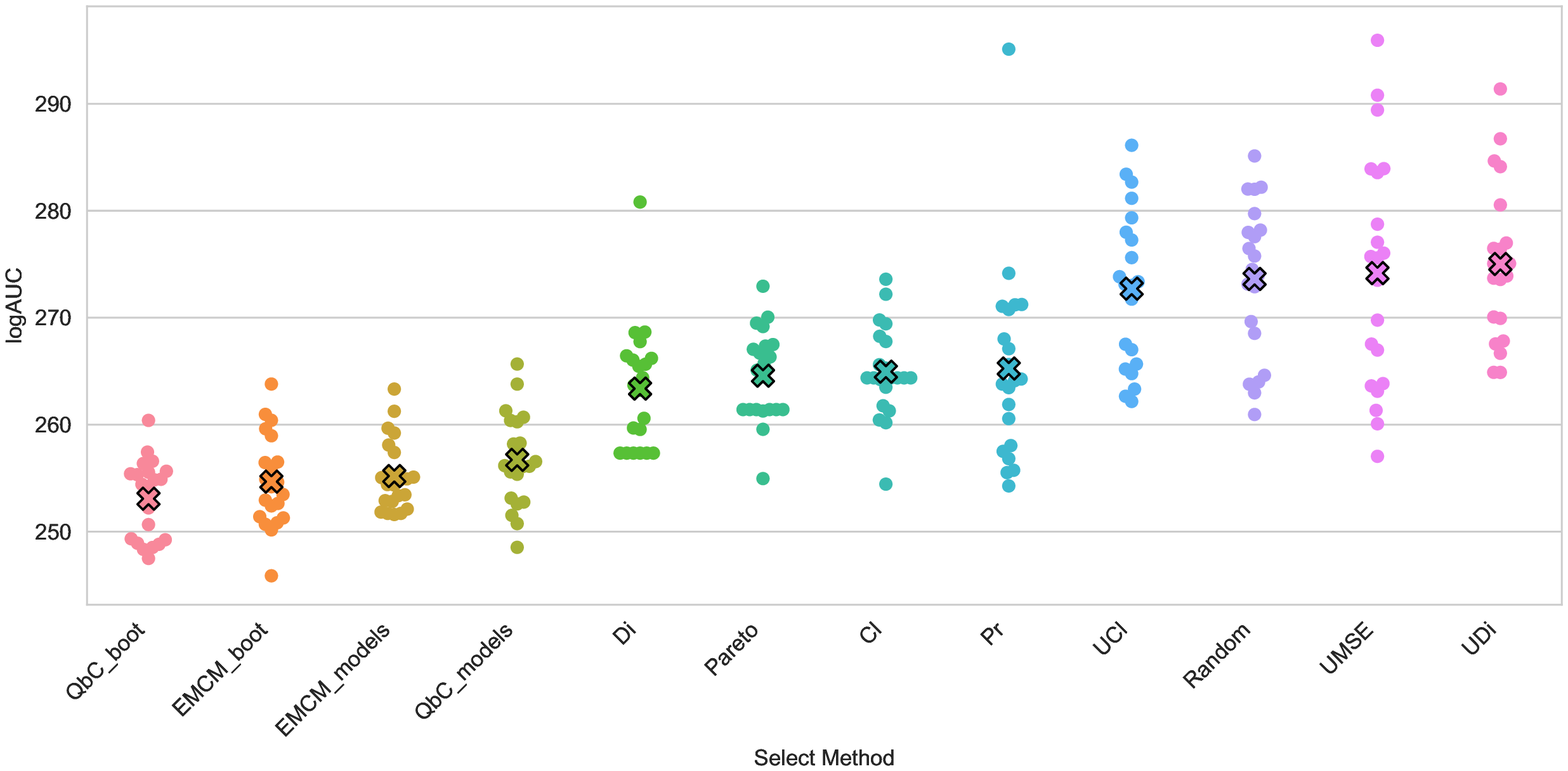}
	\caption{Comparison between selection methods (Section~\ref{sec:prelim}) using logAUC. Table~\ref{tab:quality} provides statistical significance results}
	\label{fig:selectAUC}
\end{figure}

Figure~\ref{fig:selectAUC} compares the performance of all (initialization and selection) methods in terms of logAUC. Each initialization method (Section~\ref{sec:initexp}) is represented using a separate dot. The non-learning (initialization) methods outperform the (learning) uncertainty-based methods. This serves as another evidence to the difficulty of adapting uncertainty methods to regression methods, suggesting the possible preference of using a heuristic approach, such as Distances or Pareto, for sample selection.  

The lowest value in the plot belongs to EMCM$_{boot}$ that uses boot choice of initialization (Section~\ref{sec:initexp}). Yet, overall, we observe that QBC$_{boot}$ performs better on average, and the difference between methods is insignificant. This, in turn, emphasizes the impact of the initialization method on the final outcome. Therefore, a possible direction for a future work involves a careful analysis of the combination of initialization and selection methods when applying active learning techniques.  
\else\fi

\subsubsection{Analyzing the Quality of Learning}\label{sec:qualityexp}
\ifdefined\TechReport
Sections~\ref{sec:regexp}-\ref{sec:selectexp} focused on MSE and logAUC as common practice measures of success for active learning~\cite{baram2004online}. Next, we examine other aspects of active learning, following the discussion in Section~\ref{sec:measures}.
\else
Next, we examine other aspects of active learning, following the discussion in Section~\ref{sec:measures}.
\fi

\ifdefined\TechReport
\begin{table*}[t]
	\centering
	\caption{Comparison among all methods (including both initialization and selection) in terms of MSE, mseAUC, logAUC, ASD, WASD, and FTC (Section~\ref{sec:measures}). Statistical significance of logAUC performance over random (based on the same initialization steps) is marked with an asterisk.}
	\label{tab:quality}
	\scalebox{1.1}{\begin{tabular}{|ll|c|c|c|c|c|c|c|c|}
		\hline
		\multicolumn{2}{|c|}{Method}                      & MSE (20) & MSE (50) & MSE (100)   & mseAUC & logAUC & ASD & WASD & FTC  \\\hline
		\multirow{3}{*}{\shortstack[l]{Initialization \\ Methods}} & Cl         & 16.39 & 10.72 & 10.25 & 1315.11                 & 261.76                     & 0.98                    & 0.34                       & 97.00                    \\
		& Di           & 17.25 & 10.10 & 9.49  & 1267.56                 & 257.34$^{*}$                     & 1.14                    & 0.42                       & 99.00                    \\
		& Pareto             & 18.33 & 10.70 & 10.22 & 1294.59                 & 259.56                     & 1.12                    & 0.46                       & 96.00                    \\\hline
		& Pr         & 17.77 & 10.52 & 9.53  & 1278.00                 & 257.50$^{*}$                     & 1.40                    & 0.54                       & 93.00                    \\\hline
		\multirow{3}{*}{\shortstack[l]{Uncertainty\\ Methods}} & UCl       & 17.54 & 11.25 & 10.56 & 1336.36                 & 263.33                     & 0.92                    & 0.31                       & 99.00                    \\
		& UDi & 20.54 & 12.14 & 10.96 & 1430.77                 & 270.07                     & 1.15                    & 0.46                       & 100.00                   \\
		& UMSE            & 15.56 & 10.89 & 9.90  & 1295.26                 & 260.08                     & 0.90                    & 0.31                       & 99.00                    \\\hline
		\multirow{2}{*}{\shortstack[l]{Expected Model\\ Change Maximization}} & EMCM$_{boot}$             & 13.01 & \textbf{8.72}  & 8.67  & \textbf{1127.48}                 & \textbf{245.87}$^{*}$                     & 0.98                    & 0.38                       & 100.00                   \\
		& EMCM$_{model}$                & 12.63 & 10.65 & 9.88  & 1186.87                 & 251.70$^{*}$                     & 0.91                    & 0.34                       & 100.00                   \\\hline
		\multirow{2}{*}{\shortstack[l]{Query by\\ Committee}} &QBC$_{boot}$              & 13.52 & 9.64  & \textbf{8.41}  & 1164.28                 & 249.33$^{*}$                     & 0.93                    & 0.35                       & 99.00                    \\
		& QBC$_{model}$                 & \textbf{12.08} & 9.30  & 9.13  & 1154.09                 & 248.53$^{*}$                     & 0.89                    & 0.29                       & 94.00                    \\\hline
		& Random                      & 17.95 & 11.11 & 10.76 & 1342.70                 & 263.78                     & \textbf{0.72}                    & \textbf{0.17}                       & \textbf{64.00} \\\hline                  
	\end{tabular}}
\end{table*}
\else
\begin{table}[t]
	\centering
	\caption{Comparison among all methods (including both initialization and selection) in terms of MSE, mseAUC, logAUC, ASD, WASD, and FTC (Section~\ref{sec:measures}). Statistical significance of logAUC performance over random (based on the same initialization steps) is marked with an asterisk.}
	\label{tab:quality}
	\scalebox{0.68}{\begin{tabular}{|l|c|c|c|c|c|c|c|c|}
			\hline
			\multicolumn{1}{|c|}{Method}                      & MSE(20) & MSE(50) & MSE(100)   & mseAUC & logAUC & ASD & WASD & FTC  \\\hline
			 Cl         & 16.39 & 10.72 & 10.25 & 1315.11                 & 261.76                     & 0.98                    & 0.34                       & 97                    \\
			 Di           & 17.25 & 10.10 & 9.49  & 1267.56                 & 257.34$^{*}$                     & 1.14                    & 0.42                       & 99                    \\
			 Pareto             & 18.33 & 10.70 & 10.22 & 1294.59                 & 259.56                     & 1.12                    & 0.46                       & 96                    \\\hline
			 Pr         & 17.77 & 10.52 & 9.53  & 1278.00                 & 257.50$^{*}$                     & 1.40                    & 0.54                       & 93                   \\\hline
			 UCl       & 17.54 & 11.25 & 10.56 & 1336.36                 & 263.33                     & 0.92                    & 0.31                       & 99                    \\
			 UDi & 20.54 & 12.14 & 10.96 & 1430.77                 & 270.07                     & 1.15                    & 0.46                       & 100                   \\
			 UMSE            & 15.56 & 10.89 & 9.90  & 1295.26                 & 260.08                     & 0.90                    & 0.31                       & 99                    \\\hline
			 EMCM$_{boot}$             & 13.01 & \textbf{8.72}  & 8.67  & \textbf{1127.48}                 & \textbf{245.87}$^{*}$                     & 0.98                    & 0.38                       & 100                   \\
			 EMCM$_{model}$                & 12.63 & 10.65 & 9.88  & 1186.87                 & 251.70$^{*}$                     & 0.91                    & 0.34                       & 100                 \\\hline
			 QBC$_{boot}$              & 13.52 & 9.64  & \textbf{8.41}  & 1164.28                 & 249.33$^{*}$                     & 0.93                    & 0.35                       & 99                    \\
			 QBC$_{model}$                 & \textbf{12.08} & 9.30  & 9.13  & 1154.09                 & 248.53$^{*}$                     & 0.89                    & 0.29                       & 94                    \\\hline
			 Random                      & 17.95 & 11.11 & 10.76 & 1342.70                 & 263.78                     & \textbf{0.72}                    & \textbf{0.17}                       & \textbf{64} \\\hline                  
	\end{tabular}}
\end{table}
\fi

Table~\ref{tab:quality} compares all (initialization and selection) methods, using the performance measures of mean squared error (MSE) (with  20, 50, and 100 iterations), area under MSE curve (mseAUC) and its logarithmic counterpart (logAUC), smoothness as determined by absolute second difference (ASD) and weighted absolute second difference (WASD), and convergence (FTC). 

\ifdefined\TechReport
In Section~\ref{sec:selectexp}, we showed QBC and EMCM methods to have superior performance, with final MSE values ranging between $8.41$ (QBC$_{boot}$) and $9.88$ (EMCM$_{model}$) and logAUC between $245.87$ (EMCM$_{boot}$) and $251.7$ (EMCM$_{model}$). In addition, all QBC and EMCM methods perform significantly better than random along the whole learning course and even with fewer training examples (the MSE of QBC and EMCM after 50 iterations is lower than the MSE of random after 100 iterations). Yet, when it comes to smoothness and convergence, random selection provides a smooth learning curve that rapidly converges ($64$th iteration compared to other methods that converge just before the finish line). This dominance, in terms of smoothness and convergence, can be explained by the inability of random to improve on its performance, smoothly converging to a poor result fast. Additionally, random was also averaged over 15 runs, which may also affect its smoothness. Interestingly, the non-random methods, UCl and UMSE are only second to QBC$_{model}$ in terms of WASD. This indicates that although using uncertainty methods for regression results in inferior performance (compared to QBC and EMCM) it does provide a smooth learning curve. Finally, all non-random methods coverage roughly at the same point towards the end of the experiment. This serves as evidence that  additional training data (either in a supervised learning setup or by extending the experiment horizon) can further improve on the reported results.     
\else
In a technical report~\cite{tech}, we showed that QBC and EMCM (Section~\ref{sec:prelim}) have superior performance (see also Table~\ref{tab:quality}), with final MSE values ranging between $8.41$ (QBC$_{boot}$) and $9.88$ (EMCM$_{model}$) and logAUC between $245.87$ (EMCM$_{boot}$) and $251.7$ (EMCM$_{model}$). In addition, all QBC and EMCM methods perform significantly better than random. Yet, when it comes to smoothness and convergence, random selection provides a smooth learning curve that rapidly converges ($64$-th iteration compared to other methods that converge just before the finish line). This dominance, in terms of smoothness and convergence, can be explained by the inability of random to improve on its performance, smoothly and quickly converging to a poor result. Additionally, random was also averaged over 15 runs, which may also affect its smoothness. Interestingly, the non-random methods, UCl and UMSE are only second to QBC$_{model}$ in terms of WASD. This indicates that although using uncertainty methods for regression results in inferior performance (compared to QBC and EMCM) it does provide a smooth learning curve. Finally, all non-random methods converge roughly at the same point towards the end of the experiment. This indicates that additional training data (either in a supervised learning setup or by extending the experiment horizon) can further improve reported results.     
\fi    


\subsubsection{Feature Analysis}\label{sec:featuresexp}
%

\begin{figure}[h]
	\centering
	\includegraphics[width=.75\linewidth, height=.55\linewidth]{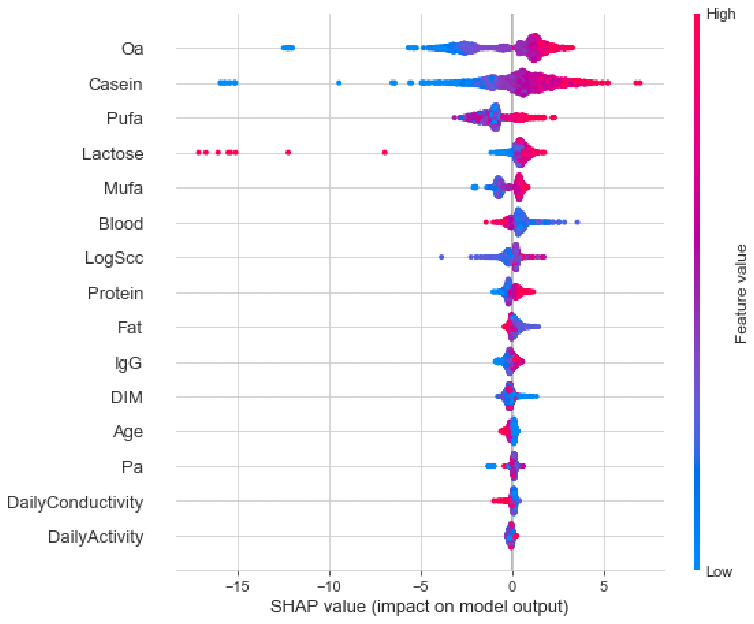}
	\caption{Feature importance using SHAP}
	\label{fig:features}
\end{figure}

\add{In our final analysis we recall one of the core goals of this work, namely understanding the impact of cow and milk features on cottage cheese production.} To do so, we provide feature importance analysis using SHAP~\cite{shap}, a state-of-the-art technique to interpret machine learning models. The top 15 important features are given in Figure~\ref{fig:features} ranked by their shap values, representing their relative importance in the trained model.

The most dominant feature is $Oa$, which together with $Pufa$ (3rd) and $Mufa$ (5th), represent the dynamics of milk coagulation process (see Example~\ref{ex:sensors}) and higher values affect CF positively. This serves as a new insight to the milk industry as no prior research has considered these properties as quality predictors. Milk enzymatic coagulation ($Casein$ 2-nd), conductivity, and lactose (4-th) validate our analysis as they are known to have an effect on quality~\cite{kappa93,Leitner2008,Katz2016}. Here, $Casein$ is more significant and higher values have a positive impact on CF. Note that, although lactose mostly has a positive effect, there are a few outliers with opposite effect.  


\section{Related Work}
\label{sec:related}

There are numerous machine (and deep) learning solutions for the food industry domain~\cite{jimenez2019alternative,li2020application,yoon2020delivering} and other domains with similar characteristics, such as transportation~\cite{gal2017traveling,petersen2019multi} and air pollution~\cite{qi2018deep,haehnel2020using}. Common to all of these works is the availability of labeled data. We focus on a setting where labeled data is scarce, applying active learning to efficiently utilize available data samples.

Active learning methods prioritize data to be labeled in order to maximize the models ability to learn. Active learning is widely used for classification problems~\cite{settles1995active,lang1995newsweeder,cai2013maximizing,kholghi2015external} and was also used for several data management tasks. Arasu \emph{at el.}~\cite{arasu2010active} consider the task of record matching packages in an active learning setting and Mozafari \emph{at el.}~\cite{mozafari2014scaling} use active learning for an optimization strategy in crowd-sourced databases. More recently, the tasks of interactive database exploration~\cite{huang12optimization}, time series and text classification~\cite{liang2020active,cormack2016scalability}, enhancing database systems~\cite{ma2020active} have also been addressed with active learning techniques. In this work we use active learning for regression, which, has yet to gain wider popularity.

Aiming to use active learning for regression, Burbidge \emph{et al.} suggests using a committee of classifiers (or regressors) on different subsets of the labeled data~\cite{burbidge2007active}. Cai \emph{et al.} presents a data sampling solution in the context of regression, which queries the examples that lead to the largest model change~\cite{cai2013maximizing}. Finally, O'Neill compares active learning techniques in the context of linear regression models~\cite{o2015evaluation}. In this work, we compare the aforementioned methods and others in our unique scenario of the milk industry.

Consensus is not yet apparent regarding the evaluation of active learning methods. Several works base their analysis on qualitative criteria~\cite{shen2004multi,wu2018pool,huang2010active}. Others focus on quantitative evaluation~\cite{settles1995active}. Specifically, Baram \emph{et al.} suggests to evaluate the performance of an active learner over time using the AUC measure~\cite{baram2004online}. In our analysis, we use the logAUC as our comparison measure among methods. We also adapt additional evaluation measures, \emph{e.g.}, ASD and FTC, to account for additional perspectives of active learning. 

Finally, several previous works analyze cheese-milk quality~\cite{kappa93,Leitner2008,Katz2016}. Leitner \emph{et al.} investigates the effect of milk storage on quality using individual cows. Katz \emph{et al.} examines the importance of milk composition for the final product~\cite{Katz2016}. In our empirical evaluation we validate some of the findings of these studies, enriching the literature with new insights that demonstrate the dominance of dynamics of milk coagulation process as quality predictors.
\section{Conclusions} \label{sec:conclusions}

This work focuses on the problem of acquiring a training dataset with limited resources. We offer an active learning data collection schedule methodology to predict production quality with limited amount of annotated raw material data. A real-world use-case in the milk industry demonstrates the approach usefulness, using milk and cow features to predict the quality of cottage cheese. 

Our suggested algorithmic solution is composed of two main phases handling varying sizes of available training data. First, with a small training set, we utilize human meta-knowledge to identify useful samples. Then, as the size of labeled training data set increases, we extend state-of-the-art methods of active learning to handle regression-based problems.  

An extensive empirical evaluation that uses data collected at Afimilk laboratory demonstrates the effectiveness of our proposed approach. The results indicate that when the training set is limited, random forest regressor preforms better than linear, Multi-layer Perceptron neural network, and XGBoost regressors, the active learning benefits from an initialization phase, and committee-based selection (expected model change maximization and query by committee) outperforms other selection methods. Feature analysis reveals new insights, indicating that the dynamics of milk coagulation process impact the production quality. 

\add{The suggested active learning framework was developed for (and based on) a real-world setting of a cheese production line. The presented framework is applicable to any IoT scenario where: 1) labeled data is scarce and hard or expensive to obtain, 2) prediction is needed for a process outcome (using regression rather than classification), and 3) the pool of samples continuously changes. As part of our ongoing research we aim to experiment with additional industrial domains such as viticulture.}

\add{In future work we also intend to analyze the active learning methodologies in additional settings. This includes combining various active learning methods in a single experiment, analyze the impact of the budget and experiment horizon, taking advantage of the relationship between available sample sets (\emph{e.g.,} same cow at different timestamps), and designing a domain specific selection methodology directly aiming at maximizing one or more of the suggested quality measures. In addition, in our experiments we used individual cow samples. We aim to extend our framework to handle varying granularity levels, including milk coming from different cows in the same shed (containing several indistinguishable cows) and cows from multiple farms.}


\bibliographystyle{ACM-Reference-Format}
\bibliography{refs}

\end{document}